%% file: acl_latex.tex
\pdfoutput=1

\documentclass[11pt]{article}

\usepackage[preprint]{acl}

\usepackage{times}
\usepackage{latexsym}

\usepackage[T1]{fontenc}

\usepackage[utf8]{inputenc}

\usepackage{microtype}

\usepackage{inconsolata}

\usepackage{graphicx}

\usepackage{bm}
\usepackage{amsmath}
\usepackage{xspace}
\usepackage{float}
\usepackage{graphicx}
\usepackage{multirow}
\usepackage{booktabs}
\usepackage{paralist}

\newcommand{\ie}{\emph{i.e.,}\xspace}
\newcommand{\eg}{\emph{e.g.,}\xspace}

\newcommand{\indexmain}[2]{\texttt{#1}$\rightarrow$\texttt{#2}}

\title{Language Models are Universal Embedders}

\author{
Xin Zhang\textsuperscript{1,2}, Zehan Li, Yanzhao Zhang, Dingkun Long \\
\textbf{Pengjun Xie, Meishan Zhang\textsuperscript{1}\thanks{Correspondence. Code: \url{github.com/izhx/uni-rep}}, Min Zhang\textsuperscript{1}} \\
\textsuperscript{1}Harbin Institute of Technology, Shenzhen, \textsuperscript{2}The Hong Kong Polytechnic University \\
\texttt{zhangxin2023@stu.hit.edu.cn \{zhangmeishan,zhangmin2021\}@hit.edu.cn}
}

\begin{document}
\maketitle
\begin{abstract}
In the large language model (LLM) revolution, embedding is a key component of various systems, such as retrieving knowledge or memories for LLMs or building content moderation filters.
As such cases span from English to other natural or programming languages, from retrieval to classification and beyond,
it is advantageous to build a unified embedding model rather than dedicated ones for each scenario.
In this context, the pre-trained multilingual decoder-only large language models, \eg BLOOM, emerge as a viable backbone option.
To assess their potential, we propose straightforward strategies for constructing embedders and introduce a universal evaluation benchmark.
Experimental results show that our trained model is proficient at generating good embeddings across languages and tasks, even extending to languages and tasks for which no finetuning/pretraining data is available.
We also present detailed analyses and additional evaluations.
We hope that this work could encourage the development of more robust open-source universal embedders.
\end{abstract}

\section{Introduction}
\label{sec:intro}

Embeddings, which transform discrete text or code sequences into continuous vectors, are widely used in many fields \citep{li2022brief,Neelakantan2022TextAC}.
They have recently gained broader attention by manipulating knowledge and memories for large language models (LLMs) and LLM-based agents \citep{peng2023check,song2022llm,wang2023voyager}.
In such scenarios, their usages are inevitably coupled with different languages and tasks.
This brings a demand for robust and universal embedders, where one single model can be applied across diverse tasks and languages, encompassing both natural and programming languages.

The common approach to building effective embedders is finetuning pretrained language models through contrastive learning on pairs of sentences \citep{Neelakantan2022TextAC,wang2022text}.
In practice, BERT-style pretrained transformer encoders are de facto standard choices, deriving powerful open-source models like E5 \citep{wang2022text}, BGE \citep{xiao2023c} and GTE \citep{li2023towards}.
However, these encoders have encountered difficulties in constructing universal embeddings because there are currently no available encoders that simultaneously support multiple natural languages and programming languages.

A possible solution is to use multilingual large language models (mLLM), such as BLOOM \citep{Scao2022BLOOMA1} series.
These models adopt a decoder-only architecture and are pretrained on meticulously curated, large-scale, multilingual corpora, ROOTS \citep{laurenccon2022bigscience}, by the next token prediction objective.
They are not only skilled in English but also excel in other languages, including natural ones such as Chinese and programming languages like Python, showing their wide-ranging language abilities.

\begin{figure}
\centering
\includegraphics[width=\columnwidth]{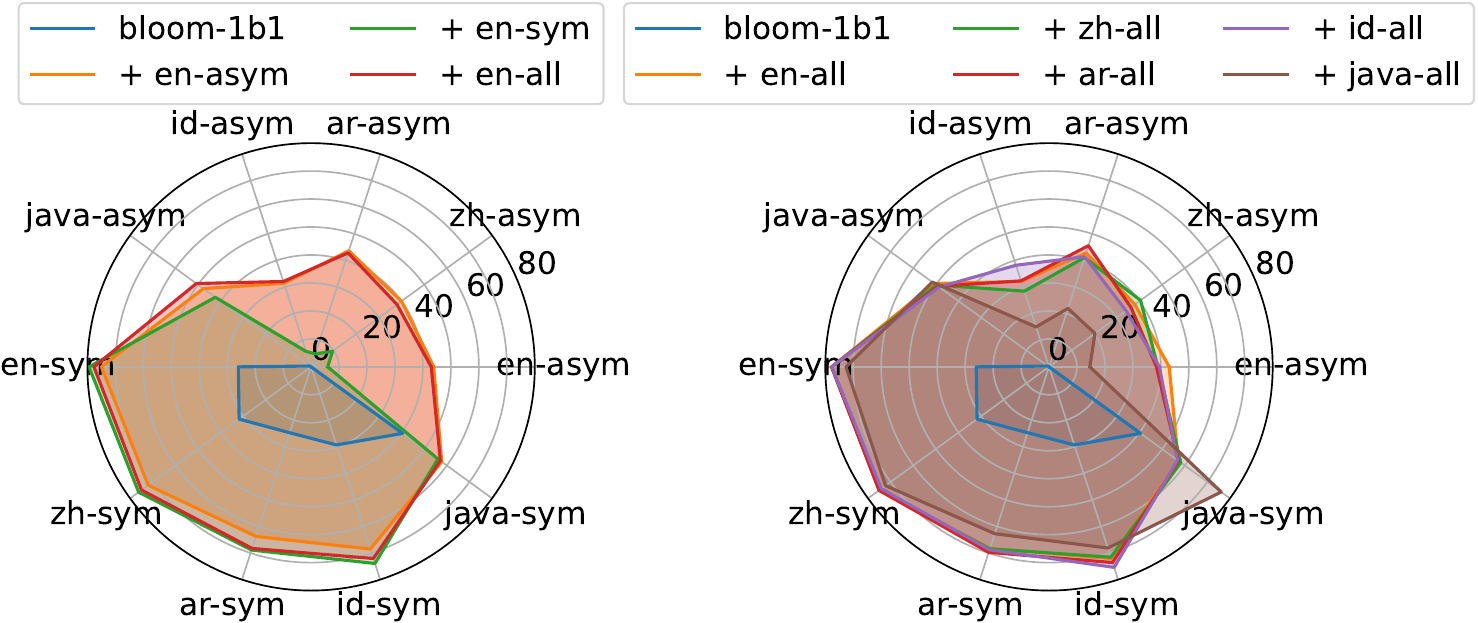}
\caption{
The performance comparison of finetuned BLOOM models on our compiled universal embedding benchmark, details refer to Table \ref{tab:main}.
}
\label{fig:radar}
\end{figure}

Therefore, one major question arises: \emph{is it feasible to derive universal embedders from mLLMs?}
To study this inquiry, we examine two scenarios:
(1) \textbf{Task versatility}: we explore strategies of data compositions that enable the model to adapt effectively to a variety of embedding tasks.
(2) \textbf{Multilinguality}: we investigate the process of obtaining embeddings across multiple languages using limited data, especially considering that some of them are hard to acquire suitable training data.
By synthesizing insights from above cases, we evaluate whether mLLMs can be trained to generate high-quality embeddings across both languages and tasks.

In practice, we construct embedders by conventional methods (detailed in \S\ref{sec:model}) based on BLOOM \citep{Scao2022BLOOMA1} models.\footnote{
Recently released Qwen1.5 is another viable option, we list the experiments in Appendix \ref{sec:qwen}.
}
For task versatility, in line with prior works \cite{wang2022text,muennighoff2022sgpt}, we categorize all embedding tasks into symmetric and asymmetric types and combine datasets from both sides for training (\S\ref{sec:data}).
Regarding multilinguality, we employ parameter-efficient fine-tuning to maximally preserve the modeling abilities of various languages (\S\ref{sec:peft}).
For evaluation, we select 5 languages (4 natural, 1 programming) and compile a universal embedding benchmark (\S\ref{sec:exp-design}).
All models are trained with monolingual data and evaluated on the benchmark (as shown in Figure \ref{fig:radar}), which helps us to analyze the performance of different languages, \eg densely, lessly or not pretrained ones, more effectively.

Through extensive experiments, we find that:
\begin{compactitem}
\item Combining datasets of both symmetric and asymmetric types can achieve task versatility across languages.
\item For pretrained languages, mLLMs can provide high-quality embeddings, even when fine-tuning occurs with data exclusively from other languages.
\item mLLMs show some extent generalizations to languages that are not pretrained, and the performance can be greatly improved by finetuning on data of these unseen languages.
\end{compactitem}
We believe that \emph{mLLMs are feasible and show great potential in building universal embedders.}

Additionally, we provide various detailed analyses (\S\ref{sec:analysis}, \S\ref{sec:scaling-ablation}, \S\ref{sec:extend}), \eg scaling the model size, and the model performance in additional benchmarks such as MTEB \citep{muennighoff-etal-2023-mteb} and CodeSearchNet \citep{husain2019codesearchnet}, to better understand the model behaviors.
We hope that our findings could foster the development and research of more powerful universal embedders.

\section{Method}

\begin{figure}
\centering
\includegraphics[width=\columnwidth]{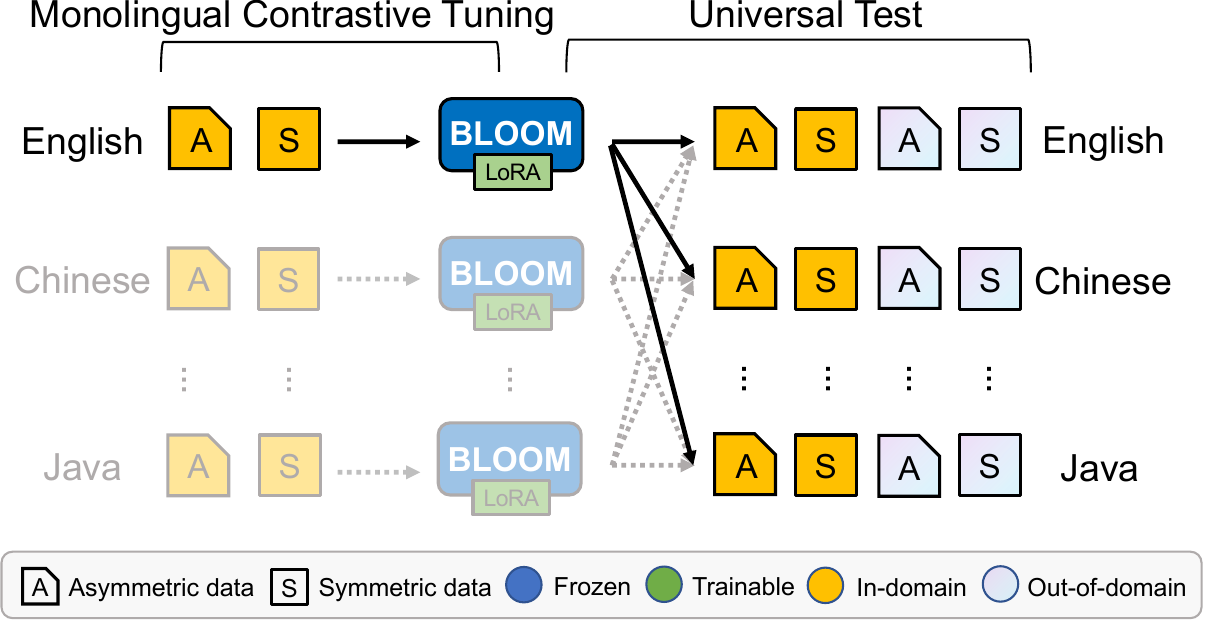}
\caption{
The outline of our main evaluation process.
We finetune BLOOM to generate embeddings by [EOS] with contrastive loss on monolingual data, and analyze performance by multilingual tests from various tasks.
The solid lines in the graph show English as an example.
}
\label{fig:method}
\end{figure}

Figure \ref{fig:method} shows our method and evaluation.
For clarity, the details of embedding model are not presented.
Next, we describe this model design.

\subsection{Embedding Model}
\label{sec:model}

Our model design mainly follows the standard practice of previous work \citep{muennighoff2022sgpt,Neelakantan2022TextAC}.
Given a text or code input $x$, we append special tokens, [BOS]$_t$ and [EOS]$_t$, to the start and end of $x$ respectively, where $t$ represents the input type.\footnote{
We set two input types, \ie query and document.
If not specified, the input is encoded as \emph{query} by default. We only use the \emph{document} type in retrieval tasks.
}
We take the last token state from the model output, \ie the representation of [EOS]$_t$, as the embedding $\bm{e}$ of the input text $x$.

The contrastive learning objective involves positive and hard-negative examples \citep{reimers-gurevych-2019-sentence}.
For each positive pair ($x$, $x^+$) in trainset, where $x^+$ is the sequence similar or relevant to $x$, we build the training instance \{$x$, $x^+$, $x^-_1$, \dots,  $x^-_N$\} with $N$ negative examples $x^-$ from the data (\S\ref{sec:data}).
We optimize the InfoNCE \citep{pmlr-v119-chen20j} contrastive loss:
\begin{equation} \label{eq:loss} \small
\mathcal{L} = - \log \frac{ \text{exp}(f_\theta(x, x^+)) }{
\text{exp}(f_\theta(x, x^+)) + \sum_{j=1}^{N}\text{exp}(f_\theta(x, x^-_i))
}
\end{equation}
where $f_\theta(x, y) = \text{cos}(\bm{e}_x, \bm{e}_y) / \tau$ denotes the function that computes the cosine similarity between two embeddings $\bm{e}_x$, $\bm{e}_y$ of inputs $x$, $y$ parameterized by $\theta$ of the model.
$\tau$ is the temperature hyperparameter which is set to $0.05$ in our experiments.

\subsection{Parameter Efficient Fine-Tuning for Multilinguality}
\label{sec:peft}

In finetuning, extensive parameter optimization can lead to catastrophic forgetting, causing models to lose their ability to model languages not included in the fine-tuning data \citep{mao2022less}.
This is a significant concern, especially for languages where paired data for contrastive learning are scarce.
In such cases, we depend on the inherent capability of model to acquire qualified embeddings, making the prevention of catastrophic forgetting essential to maintain multilingual performance.

Parameter Efficient Fine-Tuning presents a solution to balance these two aspects \citep{badola-etal-2023-parameter}, which enhances performance on target tasks while limit the updates to parameters.
Therefore, we employ it to maximize multilingual performance, focusing on popular methods like Bitfit \citep{ben-zaken-etal-2022-bitfit} and LoRA \citep{hu2021lora}.
In order to explore the model potential as much as possible, we use data from a single language in finetuning, which has demonstrated strong competitiveness \citep{wang-etal-2022-english}.

\begin{table}
\setlength{\tabcolsep}{4pt}
\resizebox{\columnwidth}{!}{
\tabcolsep=0.105cm
\begin{tabular}{c|cc|cc}
\toprule[1pt]
Language & Asymmetric & \#train & Symmetric & \#train \\
\midrule
Natural & mMarco-google & 499,184 & SNLI + MNLI & 281,230 \\
Java & CodeSearchNet & 454,451 & BigCloneBench & 450,862\\
\bottomrule[1pt]
\end{tabular}}
\caption{
Statistics of training data used in each language.
The SNLI+MNLI is translated to other languages by GPT-3.5-turbo API.
}
\label{tab:data}
\end{table}

\subsection{Data Composition for Task Versatility}
\label{sec:data}

Downstream embedding tasks can be categorized into two types: symmetric and asymmetric \citep{wang2022text,sy-etal-2023-oneemb}.
To ensure the versatility, we use both types data (Table \ref{tab:data}).

\paragraph{Asymmetric Data}
Query-to-passage/document retrieval is a typical asymmetric embedding task, focusing on capturing semantic relevance between texts \citep{muennighoff2022sgpt}.
The model is trained to maximize the similarity of vectors between a query and its most relevant candidate.
Consistent with previous studies, we select the MSMARCO passage ranking \citep{nguyen2016msmarco} and its translated version mMARCO \citep{bonifacio2021mmarco}.


\paragraph{Symmetric Data}
Natural language inference is an exemplary symmetric task that aligns well with the requirements of contrastive learning, where the semantic similarity between texts is gauged based on the similarity of their embeddings.
The training instances comprise sentences with at least one entailment (positive) and one contradiction (negative).
We utilize two classic English datasets, \ie SNLI \citep{bowman-etal-2015-large} and MNLI \citep{williams-etal-2018-broad}, and translate them into other languages.

For programming languages, clone detection focuses on the similarity between codes, where BigCloneBench \citep{svajlenko-2014-bcb} is used as the symmetric.
However, it is hard to find a suitable dataset that measures code to code relevance\footnote{
\citet{sedykh2023searching} introduced a code-to-code search dataset based on StackOverflow but it is not public yet.
}.
As a compromise, we use CodeSearchNet \citep{husain2019codesearchnet} which match codes and their comments.


\input{tables/main2.tex}

\section{Main Experiments}


To assess the viability of converting mLLMs into universal embedding models, we conduct two parts of experiment.
The first part aims to evaluate the potential of the LMs and validate employed strategies on the compiled benchmark (\S\ref{sec:exp-design}).
We expand to broader open evaluations in the second part (\S\ref{sec:extend}).


\subsection{Design of Controlled Experiments}\label{sec:exp-design}
The universal embedding encompasses two dimensions: (1) multilingual, including both natural and programming languages; (2) multitask, addressing both symmetric and asymmetric embedding tasks.
Conducting comprehensive evaluations and analyses can be quite complex and challenging, given the significant variations in task scope and difficulty across different languages.
Therefore, to facilitate research and comparison, we initially focus our experiments on a limited set of languages and tasks.

\paragraph{Evaluation benchmarks.}
For both symmetric and asymmetric task categories, we select two benchmarks each.
One is in-domain, which is the corresponding evaluation of training data.
For the asymmetric (resp. symmetric) part of natural languages, it is devset of mMarco (resp. testset of STS Benchmark \footnote{
The STS-B data are originated from SNLI. We use the translated version from \url{hf.co/datasets/stsb_multi_mt} .
} \citep{cer-etal-2017-semeval}).
The other is an out-of-domain evaluation, which is MIRACL multilingual retrieval \citep{zhang2022making} devset (resp. MASSIVE \citep{fitzgerald2022massive} testset) for the asymmetric (resp. symmetric) of natural languages.
The out-of-domain asymmetric (resp. symmetric) testset for code is xCodeEval/nl-code-search \citep{khan2023xcodeeval} (resp. GoogleCodeJam \citep{zhao2018deepsim}).

\paragraph{Evaluation languages.}
Java is only one choice for code experiments as the training and evaluation data are hard to find for other languages.
For natural ones, we list all languages shared by mMarco, MIRACL and BLOOM pretraining in Table \ref{tab:lang-brief}.
We select English, Chinese, Arabic and Indonesian for main experiments as they are from different language families and with different ratio in ROOTS.

\paragraph{Implementation details.}\label{sec:impl}

We finetune BLOOM models by LoRA \citep{hu2021lora} with r of $64$.
We append special tokens to the vocabulary, initialize their embeddings randomly, and update them as well.
We use AdamW optimizer with learning rate (lr) $5e$-$5$ and a cosine learning rate schedule, with warmup of $10\%$ steps, and decay final lr down to $10\%$ of the peak lr.
We use GradCache \citep{gao-etal-2021-scaling} to scale up the batch size to 1024 for the \texttt{all} that combine both asymmetric and symmetric data. And that of \texttt{asym} and \texttt{sym} is 512 to keep similar optimization steps.
For each instance, we sample 7 negative examples from the hard negatives.\footnote{
Since most examples from NLI datasets have only one contradiction sentence as the hard negative, we randomly sample $6$ sentences to serve as the negative.
}
All training are conducted on 8 A100-80GB GPUs in BF16 with FlashAttention2 \citep{dao-2024-flash}.

\subsection{Results}\label{sec:main}
Table \ref{tab:main} shows the results of controlled experiments.
It is intuitive that, for each setting in every language, the in-domain trained models consistently perform the best (except the symmetric Java evaluation). 
Referencing these scores (on the diagonal), we explore the potential of Multilingual LM on the unified embeddings.
For simplicity, we index the table by a \{train (row) $\rightarrow$ eval (column)\} format, \eg \indexmain{asym-en}{sym-zh} is $72.00$.
We can also omit part of it to refer to a set of results.

\paragraph{Task versatility}
For each setting, we can observe that:
(1) \texttt{sym} models achieve poor results on asymmetric tasks (\indexmain{sym}{asym} are much lower than \indexmain{asym}{asym});
(2) \texttt{asym} models show comparable performance on symmetric tasks as the \texttt{sym} ones (\indexmain{asym}{sym} are close to \indexmain{sym}{sym});
(3) the \texttt{all} (\ie models trained on both types data) exhibit a slight decrease in asymmetric task (\indexmain{all}{asym} are slightly lower than \indexmain{asym}{asym}), but symmetric performance is improved (\indexmain{all}{sym} are better than \indexmain{asym}{sym}), resulting in the best overall score (\indexmain{all}{all} are higher than \indexmain{asym/sym}{all}).
In all (natural and programming) languages, combining symmetric and asymmetric data improves task generalization, demonstrating that \textbf{task versatility can be achieved across languages}.

\paragraph{Multilinguality}
Focusing on \indexmain{all}{all}, lower right part of Table \ref{tab:main}, we have:
(1) on the column view, for one language, the performance from other languages (except Java) trained models are close to each other and reasonably less than that of this language;
(2) on the row view, the averaged scores for each language trained models (except Java) are also similar.
On \indexmain{all}{sym}, we can also consider the above two statements to be valid with Java.
The models are not only performant in the source language, but also effective in others.
It indicates that \textbf{we can train mLLM to generate good embeddings for a language without paired data}.

\paragraph{Exception on Java}
The exception results of Java could be possibly attributed to the unsatisfactory training data.
First, the asymmetric data, \ie CodeSearchNet, is easier than mMARCO.
On asymmetric Java evaluation, natural language models could achieve comparable results to the asym-java model, but, on asymmetric natural language evaluations, the latter is substantially weaker than the former.
Thus, hard-pairs of asymmetric data would be beneficial.
Second, the symmetric data (BigCloneBench) seem to be insufficient as it is limited to only a few hundred contest problems, which is smaller than the tens of thousands of semantic groups in NLI data. A wide-coverage large-scale dataset might be helpful.

\subsection{Analysis}\label{sec:analysis}
In this subsection, we further analyze multilingual performance and mechanism.

\input{tables/lang-asym.tex}

\begin{figure}
\centering
\includegraphics[width=\columnwidth]{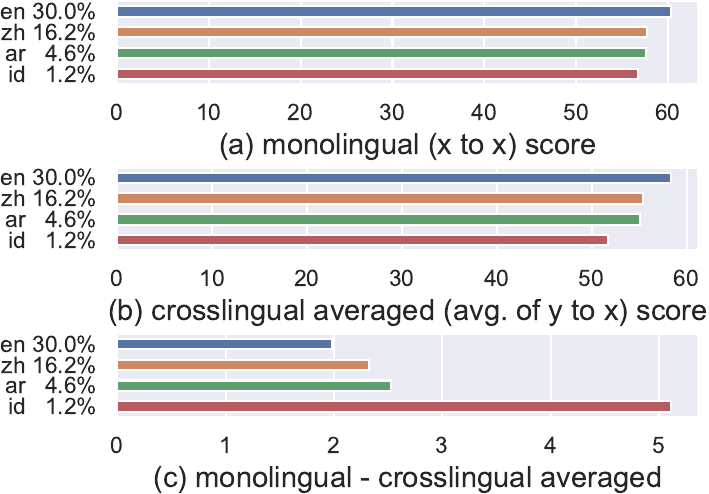}
\caption{
The plot of monolingual score (a), crosslingual averaged score (b), and their difference (c) of natural language evaluations on \indexmain{all}{all} setting.
The lower the ratio of a language in pre-training, the lower its performance, and the more significant the improvement brought by training data.
}
\label{fig:lang-ratio}
\end{figure}

\paragraph{How language pretraining ratio affect performance?} 
To explore the relationship between the performance of each language and its pretraining ratio in mLLM, we focus on natural languages in \indexmain{all}{all} setting and present the monolingual performance, cross-lingual average performance, and the differences between them in Figure \ref{fig:lang-ratio}.
From English to Indonesian, we observe decreases in both monolingual and cross-lingual performance as well as 
an increase in their difference, indicating that models have poorer representation capabilities for language with lower pretraining ratios and larger gaps to rich-pretraining languages, regardless of whether fine-tuning is applied or not.

\paragraph{Can model generalize to not pretrained languages?} 
The BLOOM models are not pretrained with some commonly used languages such as German and Japanese.
To investigate such scenario, we extend to more languages 
and focus on the \indexmain{asym}{asym} setting.
Table \ref{tab:lang-asym} displays the results of three languages that are not covered by ROOTS, \ie German (de), Russian (ru) and Japanese (ja).
First, the models trained on pretrained languages (\eg \texttt{en}) are capable on them  (\eg, \indexmain{en}{de} has a small gap with \indexmain{de}{de}).
Second, for an unpretrained language, with its fine-tuning data, mLLM not only exhibits excellent performance in this language itself but also acquires a certain level of multilingual embedding ability (it also achieves considerable scores on other languages).
Overall, mLLM achieves promising generalization.

\paragraph{Does performance correlate to language families?} 
It is also interesting to investigate whether there is a connection between language family and performance.
Focusing rows of three Indo-European languages (en, fr, es) and one Sino-Tibetan language (zh) in Table \ref{tab:lang-asym}.
The results show that the models trained on Indo-European languages indeed exhibit similar performance trends, while the model trained on zh shows significant differences on es, fr and ar, which indicates that the language family is one potential factor.
We also provide a better visualization of the results in Appendix Figure \ref{fig:lang-family} .

\begin{figure}
\centering
\includegraphics[width=\columnwidth]{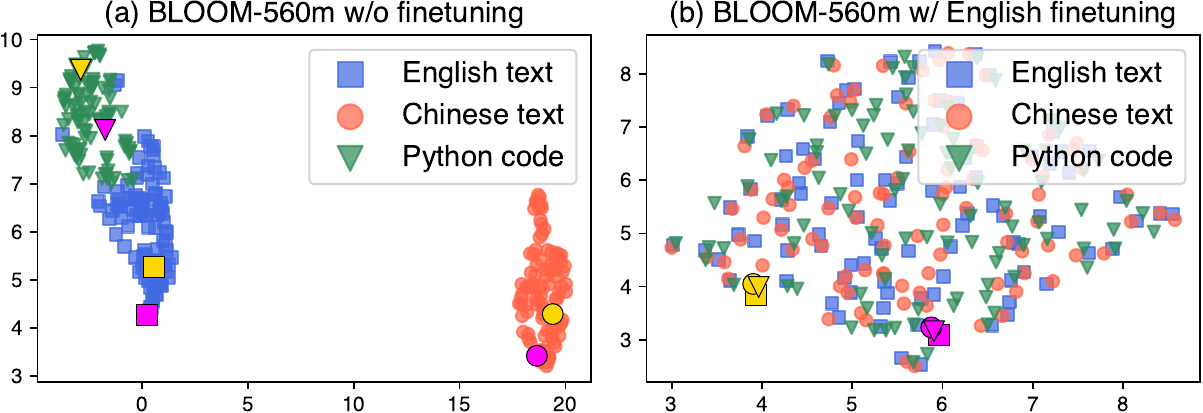}
\caption{
Visualization of 100 examples from CodeSearchNet Python, where Chinese texts are translated by GPT-3.5-turbo.
Gold and pink markers represent parallel sequences in different languages.
Before finetuning, (a), embeddings are separated by language, especially English and Chinese. After English finetuning, (b), the parallel sequences are well aligned to each other.
}
\label{fig:vis}
\end{figure}

\input{tables/scaling.tex}
\input{tables/mteb-en.tex}

\paragraph{What contributes to the multilinguality?} 
To explore why monolingual fine-tuning can also lead to satisfactory performance in other languages, we visualize the embeddings before and after fine-tuning using umap \citep{mcinnes2018umap}.
We select the top 100 text-code pairs from the CodeSearchNet test set, translate the text into Chinese, and obtain embeddings using the model trained on English.
As shown in Figure \ref{fig:vis}, before finetuning, the embeddings of each language are distributed separately.
After finetuning, all embeddings are distributed according to semantics (the text-code pair and Chinese translation are clustered together).
This indicates that monolingual contrastive learning align embeddings in the shared semantic space across languages, thereby improving performance in other languages, consistent with the finding of \citet{wang-etal-2022-english}.

\subsection{Scaling and Ablation on English}\label{sec:scaling-ablation}
In this subsection, we take English data as an example to explore scaling and ablation of LoRA.

\paragraph{Scaling model size}
All previous experiments are conducted on BLOOM-1b1.
Here, we extend the experiments to the 3b and 7b1 models.
As shown in Table \ref{tab:scaling}, the performance gradually increases as model size increases.
Additionally, for a language, the smaller the pre-training ratio, the greater the improvement brought about by scaling.

\paragraph{LoRA \textit{v.s.} full parameter tuning}
The impact of data combination has been reflected in Table \ref{tab:main}.
Now we conduct the ablation of LoRA by comparing with the full-parameter finetuned model.
In Table \ref{tab:scaling}, although full parameter fine-tuning resulted in performance improvement in English, Chinese, and Arabic, it shows a decrease in Indonesian and Java, two languages with smaller proportions of pre-training.
To ensure better performance across multiple languages, we opt for LoRA.

\section{Extended Evaluations}\label{sec:extend}
The second part experiment consists of evaluations on more tasks and domains (\S\ref{sec:domain-task-eval}), as well as diverse languages of multilingual (\S\ref{sec:multilingual-eval}) and cross-lingual (\S\ref{sec:crosslingual-eval}) tests.
We evaluate BLOOM models (1b1, 3b, 7b1) finetuned on English data. 

\subsection{Task and Domain Evaluation}\label{sec:domain-task-eval}

\paragraph{Our method improves task generalization. } 
The MTEB benchmark \citep{muennighoff-etal-2023-mteb} compiles a variety of embedding datasets for different tasks and domains.
We evaluate the generalization on MTEB English subset, which is currently one of the most comprehensive benchmark for English embeddings.
Table \ref{tab:mteb-en-results} shows the results of the English MTEB.
Compared to decoder-only models trained only on asymmetric data (SGPT series), our model significantly improves the performance on symmetric tasks (classification, clustering, STS).
We acknowledge that there is still room to go compared to the best models, which are densely trained on diverse datasets.
As our goal is to build a unified model for various languages, the score on English is already competitive enough.

\paragraph{mLLM can generalize to unseen domains.} 
To assess the domain generalization, we focus on a more challenging scenario, a Chinese multi-domain retrieval benchmark \citep{long2022multicpr} which has nearly no overlap with the training and finetuning data.
Table \ref{tab:multicpr-results} presents the results.
Our model is on par with the in-domain continue pre-trained and finetuned model \citep{karpukhin-etal-2020-dense} (\texttt{DPR-2}), which highlights the remarkable domain generalization ability of mLLM.

\input{tables/multicpr.tex}

\subsection{Multilingual Evaluation}\label{sec:multilingual-eval}

\paragraph{mLLM outperforms supervised code models.}
In main experiments (\S\ref{sec:main}), Java is the only programming language evaluated.
Now we expand the evaluations to all languages in CodeSearchNet \citep{husain2019codesearchnet}, as shown in Table \ref{tab:code-search-net}.
Our models (1b1, 3b, and 7b1) are better than supervised baselines of code \citep{feng-etal-2020-codebert,guographcodebert}, demonstrating that our approach is a promising solution in building text and code unified embeddings.
In addition to python, our models has large margins to OpenAI APIs in others.
This is reasonable given their pre-training on large-scale code-text pairs.


\input{tables/codesearchnet.tex}

\paragraph{Scaling can benefit unseen languages.} 
We now extend the symmetric evaluation with languages that are not included in the BLOOM pre-training (that of the asymmetric refer to Table \ref{tab:lang-asym}).
We conduct experiments on the multilingual testset of STS-17 \citep{cer-etal-2017-semeval}.
Following the STS evaluation protocol of MTEB, we use the Spearman correlation between the cosine similarity of the sentence embeddings and the human-annotated scores (from $1$ to $5$) as the metric.
Table \ref{tab:sts17-multi} compares the results of our models with baselines.
For languages included in the BLOOM pre-training, our models are the best.
For the unseen language (marked \textit{italic}), our models do not give competitive performance.
Nonetheless, parameter scaling leads to the increase of language capabilities, resulting in improvement scores.

\input{tables/sts17-multi.tex}

\subsection{Cross-lingual Evaluation}\label{sec:crosslingual-eval}

\paragraph{Scaling aligns unseen languages with English.}  
In Table \ref{tab:sts17-multi}, it is evident that parameter scaling can enhance monolingual performance for unseen languages.
We now investigate whether this finding still holds for cross-lingual tasks and inquire whether unseen languages are aligned with English. 
We evaluate on the BUCC bi-text mining task \citep{zweigenbaum2016towards}, which aims to find parallel sentences, often translations, from two monolingual corpora (French / Chinese / German / Russian and English).
For fair comparisons, we adopt the setting and baselines of MTEB \citep{muennighoff-etal-2023-mteb}.
Table \ref{tab:bucc} shows the F1 scores on the BUCC testset.
Similar to the multilingual results, on the pre-trained language pairs (\ie \texttt{fr-en} and \texttt{zh-en}), our models are comparable with the state-of-the-art approach, \texttt{LABSE} \citep{feng-etal-2022-language}.
On the half-covered language pairs (\textit{de}\texttt{-en} and \textit{ru}\texttt{-en}), there are consistent improvements with the model size growth, demonstrating that the embedding spaces of unseen languages are aligned to that of English.
Hence, we can affirmatively answer the research question posed earlier.

\input{tables/bucc.tex}

\section{Related Work}
\label{sec:related-work}

Text and sentence embeddings are useful for many downstream tasks and applications \citep{karpukhin-etal-2020-dense,gao-callan-2021-condenser}.
Early studies start from similar ideas of word vectors \cite{hill-etal-2016-learning,lin2017structured,pagliardini-etal-2018-unsupervised}, also shift to neural networks \citep{conneau-etal-2017-supervised} then pre-trained transformers \citep{cer-etal-2018-universal,reimers-gurevych-2019-sentence,ni-etal-2022-sentence}.
The subsequent work mainly focus on using contrastive loss to supervise or improve representation learning \citep{zhang-etal-2020-unsupervised,giorgi-etal-2021-declutr,kim-etal-2021-self,gao-etal-2021-simcse,yan-etal-2021-consert,cheng2023improving}, translation augmentation \citep{wieting-etal-2020-bilingual,zhang-etal-2021-bootstrapped}, large-scale pre-training \citep{yang-etal-2021-universal,Neelakantan2022TextAC,wang2022text}, and prompt \citep{sy-etal-2023-oneemb}.
As most of them are under specific tasks, \citet{muennighoff-etal-2023-mteb} compile MTEB with diverse tasks, domains, and languages for evaluations.
Recently, embeddings have gained attention and a batch of large-scale pretrained models have emerged, such as E5 \citep{wang2022text}, BGE \citep{xiao2023c}, GTE \citep{li2023towards}, UAE \citep{li2023angle}.
Most of them are targeted to and evaluated on English, while we explore the languages beyond English.


Pre-trained transformer encoders, \ie BERT \citep{devlin-etal-2019-bert}, or that of T5 \citep{JMLR:v21:20-074} are currently the mainstream for embedding models, which are computation-effective than encoder-decoders \citep{ni-etal-2022-sentence}.
GPT-style decoder-only models \citep{radford2018improving} are promising alternatives, since they have theoretically stronger representations 
\citep{pmlr-v139-dong21a,kexuefm-9529}.
Pioneering GPT-based studies show impressive performance on both text and code \citep{Neelakantan2022TextAC}, especially for semantic search \citep{muennighoff2022sgpt}.
We continue this line, exploring the unified embeddings across multiple natural and programming languages.
A concurrent work \citep{wang2024improving} fine-tune Mistrial-7B \citep{jiang2023mistral} with data from diverse source and carefully crafted instructions, showing state-of-the-art performance on English MTEB.
Taking into account a more general scenario with various languages, we do not use complex prompts, but only a set of special symbols for asymmetric inputs.

Multi- and cross-lingual text embeddings follow the developments of English ones, from cross-lingual word embeddings \citep{ruder-et-al-2019-asurvey} to RNNs \citep{artetxe-schwenk-2019-massively} and transformers \citep{chidambaram-etal-2019-learning,yang-etal-2020-multilingual,reimers-gurevych-2020-making,feng-etal-2022-language}.
To learn models without enough supervisions, translation information \citep{artetxe-schwenk-2019-massively,chidambaram-etal-2019-learning,goswami-etal-2021-cross,feng-etal-2022-language} and multilingual pre-trained encoders \citep{reimers-gurevych-2020-making,liu-etal-2021-fast} are explorated to improve embeddings \citep{chen2024bgem3}.
However, such BERT-like multilingual encoders do not support code, which is currently one of the crucial requirements.
Therefore, we shift our focus to pre-training decoder models that can simultaneously support natural languages and programming languages, aiming to evaluate and analyze the potential of constructing universal embeddings from them.



\section{Conclusion}
We propose the development of unified embeddings models (universal embedders) for various tasks across multiple natural and programming languages based on multilingual decoder-only models.
To evaluate the potential, we present straightforward strategies to construct embedding models from them, and design a universal embedding benchmark for evaluation and analysis.
Through extensive experiments, we demonstrated the versatility of embedders constructed from mLLMs, showing their capabilities cross languages and tasks.
The models can generate reasonably good embeddings for languages that have not been finetuned or pre-trained, and the quality can be significantly improved with the corresponding fine-tuning data.
These characteristics strongly indicate the great potential of mLM for building universal embedders.
Additionally, we provide various analyses and extended evaluations to reveal the interesting properties of the model.
We hope that our work could inspire more open-source high-quality universal embedders.

\section*{Limitations}
This work suffers from three primary limitations.
Firstly, we only evaluate the BLOOM and Qwen1.5 models as they are currently the only open-source decoder-only models available for multiple natural and programming languages.
We hope that in the future, there will be more model options to consider.
Secondly, we train the model using only monolingual data.
We have chosen to focus on monolingual fine-tuning for a clearer analysis, which helps us to fully analyze the intrinsic characteristics of different languages and the performance relationships between them.
We left mixed-language training as future work.
Thirdly, there were some anomalies in the training and evaluation for the code.
We are committed to finding higher-quality data to enhance code evaluations.


\section*{Acknowledgments}
This work receives partial support from the Natural Science Foundation of China (under Grant 624B2048), ``the Fundamental Research Funds for the Central Universities'', and the Shenzhen Science and Technology Program (under Grant ZDSYS20230626091203008).

\bibliography{custom}

\appendix

\section{Appendix}
\label{sec:appendix}

\begin{table}
\setlength{\tabcolsep}{4pt}
\resizebox{\columnwidth}{!}{
\tabcolsep=0.105cm
\begin{tabular}{cc|cc|c}
\toprule[1pt]
Code & Language & Family & Subfamily & in ROOTS (\%) \\
\midrule
ar & Arabic       & Afroasiatic   & Semitic       & 4.6   \\
zh & Chinese      & Sino-Tibetan  & Sinitic       & 16.2  \\
de & German       & Indo-European & Germanic      &  -   \\
en & English      & Indo-European & Germanic      & 30.04 \\
es & Spanish      & Indo-European & Italic        & 10.8  \\
fr & French       & Indo-European & Italic        & 12.9  \\
hi & Hindi        & Indo-European & Indo-Iranian  & 0.7   \\
id & Indonesian   & Austronesian  & Malayo-Polynesian  & 1.2  \\
ja & Japanese     & Japonic & - & - \\
ru & Russian      & Indo-European & Balto-Slavic  & - \\
\bottomrule[1pt]
\end{tabular}}
\caption{
Languages shared by mMarco and MIRACL.
}
\label{tab:lang-brief}
\end{table}

\begin{figure}
\centering
\includegraphics[width=0.95\columnwidth]{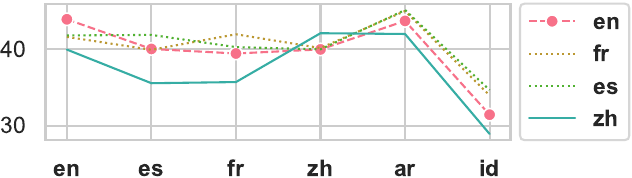}
\caption{
The plot of English (en), French (fr), Spanish (es), Chinese (zh) from Table \ref{tab:lang-asym}, where en, fr and es are all in the Indo-European family and with similar performance trends.
While the zh trained model shows differences to Indo-European ones in es, fr, and ar.
}
\label{fig:lang-family}
\end{figure}

\input{tables/main-qwen.tex}
\input{tables/mteb-en-qwen.tex}
\input{tables/ablation.tex}

\input{tables/full.tex}

\subsection{Experiments on Qwen1.5}\label{sec:qwen}
Qwen1.5 models are recently released multilingual LLMs, we conduct the main experiments on the Qwen1.5-0.5B to examine the multilingual performance (Table \ref{tab:main-qwen}) and evaluate 0.5B, 1.8B and 4B English finetuned models on MTEB English (Table \ref{tab:mteb-en-qwen}).
In Table \ref{tab:main-qwen}, Qwen1.5-0.5B is comparable to BLOOM-1b1 or even better on English (en), Chinese (zh), and Java.
But it performs poorly in Arabic (ar) and Indonesian (id).
In MTEB English, as shown in Table \ref{tab:mteb-en-qwen}, the Qwen1.5 models are significantly better than BLOOM models.

\subsection{Additional Design Analysis}\label{sec:design-analysis}

We now conduct the ablation analysis to identify the contributions of different design aspects of our approach.
We hope that this analysis can help building more robust decoder-based embedding models.
Table \ref{tab:ablation} presents the MTEB-English performance of BLOOM-560M models finetuned in different experimental settings.

\paragraph{NLI data improve symmetric tasks.}
We first investigate the effect of symmetric NLI data on different tasks.
In the line No.1 of Table \ref{tab:ablation}, we remove the NLI data and finetune the model solely using asymmetric retrieval data (MSMARCO).
Compared with our model in line No.0, the performance of classification (Class.) and STS is significantly decreased, which are typical symmetric tasks. However, these two tasks are not affected by the removal of MSMARCO data (line No.2). This demonstrates the crucial role of symmetric NLI data in achieving optimal performance in these tasks.

\paragraph{Retrieval data are irreplaceable.}
As stated above, finetuning using only NLI data (line No.2) is competitive enough for classification and STS.
However, it can not provide a satisfactory score for retrieval (Retr.), \ie $20.78$ \emph{v.s.} $40+$ of others, and also leads a drop in clustering (Clust.).
This suggests that retrieval data are crucial for building unified embedding models.

\paragraph{Multiple negatives only help retrieval.}
In line No.3 of Table \ref{tab:ablation}, we keep only one negative example in contrastive learning.
Compared to our model in line No.0, only the performance of retrieval is decreased, while other tasks have no significant change.
Considering that learning multiple negatives greatly increase the computational cost and training train, one can freely choose whether or not to use it according to the specific requirements.

\paragraph{Last special token is better representation.}
With regard to sequence encoding by decoder-based models, both \citet{Neelakantan2022TextAC} and \citet{muennighoff2022sgpt} append special tokens to the start and end of the input sequence.
On the selection of the final embedding output, \citet{Neelakantan2022TextAC} use the last special token, while \citet{muennighoff2022sgpt} use a position weighted mean pooling of the hidden states.
In line No.4 of Table \ref{tab:ablation}, we employ the weighted mean pooling on our model and observe a slight performance decrease.
Additionally, we also try to use the last special token on SGPT \citep{muennighoff2022sgpt}, achieving better average scores (line No.6) compared with the \texttt{sgpt-bloom-560m} we implemented.
Our experiments demonstrate that the last special token is more effective for unified embeddings models.

\input{tables/full-qwen.tex}

\end{document}

%% file: tables/main2.tex
\begin{table*}[tbp]
\resizebox{0.98\textwidth}{!}{
\begin{tabular}{cc|rrrrr|r|rrrrr|r|rrrrr|r}
\toprule[1.5pt]
Setting    &  Eval $\rightarrow$  & \multicolumn{6}{c|}{Asym} & \multicolumn{6}{c|}{Sym} & \multicolumn{6}{c}{All} \\
\cmidrule(lr){3-8} \cmidrule(lr){9-14} \cmidrule(lr){15-20}
Train $\downarrow$ & Lang  & \multicolumn{1}{c}{en} & \multicolumn{1}{c}{zh} & \multicolumn{1}{c}{ar} & \multicolumn{1}{c}{id} & \multicolumn{1}{c}{java} & \multicolumn{1}{|c|}{avg.} & \multicolumn{1}{c}{en} & \multicolumn{1}{c}{zh} & \multicolumn{1}{c}{ar} & \multicolumn{1}{c}{id} & \multicolumn{1}{c}{java} & \multicolumn{1}{|c|}{avg.} & \multicolumn{1}{c}{en} & \multicolumn{1}{c}{zh} & \multicolumn{1}{c}{ar} & \multicolumn{1}{c}{id} & \multicolumn{1}{c}{java} & \multicolumn{1}{|c}{avg.} \\
\midrule
\multirow{5}{*}{Asym}
& en   & \bf 43.85 & 39.93 & 43.64 & 31.43 & 47.60 & 41.29 & 75.00 & 72.00 & 63.77 & 68.51 & 57.74 & 67.40 & 59.43 & 55.96 & 53.70 & 49.97 & 52.67 & 54.35 \\
& zh   & 39.91 & \bf 42.04 & 41.94 & 28.93 & 49.24 & 40.41 & 75.05 & 72.68 & 65.32 & 68.57 & 58.54 & 68.03 & 57.48 & 57.36 & 53.63 & 48.75 & 53.89 & 54.22 \\
& ar   & 39.60 & 36.76 & \bf 46.23 & 32.70 & 50.09 & 41.08 & 75.12 & 72.82 & 65.73 & 69.85 & 56.93 & 68.09 & 57.36 & 54.79 & 55.98 & 51.27 & 53.51 & 54.58 \\
& id   & 40.00 & 35.25 & 42.19 & \bf 38.90 & 48.40 & 40.95 & 75.01 & 71.70 & 65.73 & 71.88 & 57.87 & 68.44 & 57.51 & 53.47 & 53.96 & 55.39 & 53.14 & 54.69 \\
& java & 15.36 & 19.40 & 20.44 & 13.52 & \bf 53.00 & 24.35 & 72.27 & 72.32 & 62.84 & 68.37 & 54.76 & 66.11 & 43.82 & 45.86 & 41.64 & 40.95 & 53.88 & 45.23 \\
\midrule
\multirow{5}{*}{Sym}
& en   & 5.94  & 9.46  & 4.87  & 5.80  & 42.33 & 13.68 & \bf 79.41 & 76.23 & 68.88 & 73.92 & 56.05 & 70.90 & 42.67 & 42.85 & 36.87 & 39.86 & 49.19 & 42.29 \\
& zh   & 5.15  & 7.25  & 6.76  & 6.88  & 43.13 & 13.83 & 78.84 & \bf 76.64 & 68.76 & 73.60 & 56.94 & 70.96 & 42.00 & 41.95 & 37.76 & 40.24 & 50.03 & 42.40 \\
& ar   & 5.89  & 8.19  & 8.57  & 7.38  & 42.86 & 14.58 & 78.64 & 76.01 & \bf 70.39 & 74.90 & 55.77 & 71.14 & 42.27 & 42.10 & 39.48 & 41.14 & 49.32 & 42.86 \\
& id   & 7.51  & 4.69  & 10.28 & 8.38  & 36.15 & 13.40 & 78.41 & 75.62 & 68.71 & \bf 76.17 & 54.60 & 70.70 & 42.96 & 40.16 & 39.50 & 42.28 & 45.37 & 42.05 \\
& java & 0.00  & 0.02  & 0.00  & 0.02  & 1.57  & 0.32  & 32.67 & 39.43 & 23.27 & 33.51 & 73.34 & 40.44 & 16.33 & 19.72 & 11.64 & 16.77 & 37.45 & 20.38 \\
\midrule
\multirow{5}{*}{All}
& en   & 42.97 & 37.96 & 42.85 & 32.09 & 50.70 & 41.31 & 77.65 & 74.95 & 68.26 & 72.06 & 57.14 & 70.01 & \bf 60.31 & 56.46 & 55.55 & 52.08 & 53.92 & \bf 55.66 \\
& zh   & 38.92 & 40.48 & 41.08 & 28.46 & 49.79 & 39.75 & 77.68 & 75.00 & 68.39 & 71.58 & 58.27 & 70.18 & 58.30 & \bf 57.74 & 54.73 & 50.02 & 54.03 & 54.96 \\
& ar   & 38.43 & 36.21 & 45.55 & 32.33 & 49.07 & 40.32 & 77.76 & 75.12 & 69.74 & 73.58 & 57.21 & 70.68 & 58.09 & 55.67 & \bf 57.65 & 52.95 & 53.14 & 55.50 \\
& id   & 39.48 & 34.08 & 41.41 & 38.20 & 48.58 & 40.35 & 77.69 & 74.13 & 68.78 & 75.39 & 56.82 & 70.56 & 58.58 & 54.11 & 55.09 & \bf 56.79 & 52.70 & 55.45 \\
& java & 14.62 & 20.31 & 21.97 & 15.02 & 51.56 & 24.70 & 72.60 & 72.24 & 62.74 & 68.12 & \bf 76.12 & 70.37 & 43.61 & 46.28 & 42.36 & 41.57 & \bf 63.84 & 47.53 \\
\midrule \midrule
\multicolumn{2}{c|}{Multilingual} & 43.02 & 41.69 & 46.74 & 38.73 & 49.01 & 43.84 & 77.22 & 74.88 & 69.15 & 74.66 & 60.64 & 71.31 & 60.12 & 58.28 & 57.95 & 56.70 & 54.82 & 57.57 \\
\bottomrule[1.5pt]
\end{tabular}}
\caption{
Main Results on BLOOM-1b1. The socre of the \texttt{asym} (or \texttt{sym}) is the macro average of an in-domain test and a out-of-domain test. All tests are listed in \S\ref{sec:exp-design}. The score of the \texttt{all} is the macro average of \texttt{asym} and \texttt{sym}.
}
\label{tab:main}
\end{table*}

%% file: tables/lang-asym.tex
\begin{table}[tb]
\centering
\setlength{\tabcolsep}{3pt} 
\resizebox{\columnwidth}{!}{
\begin{tabular}{c|rrrrrrrrr}
\toprule[1.5pt]
Model & \multicolumn{1}{c}{\bf en} & \multicolumn{1}{c}{\it de} & \multicolumn{1}{c}{\bf es} & \multicolumn{1}{c}{\bf fr} & \multicolumn{1}{c}{\it ru} & \multicolumn{1}{c}{\it ja} & \multicolumn{1}{c}{\bf zh} & \multicolumn{1}{c}{\bf ar} & \multicolumn{1}{c}{\bf id} \\
\midrule
\bf en  & \bf 43.85 & 19.40 & 39.99 & 39.40 & 17.53 & 27.06 & 39.93 & 43.64 & 31.43 \\
\it de  & 39.53 & \bf 35.08 & 36.70 & 36.50 & 21.31 & 29.10 & 36.93 & 41.87 & 31.66 \\
\bf es  & 41.75 & 20.88 & \bf 41.82 & 40.23 & 18.50 & 26.92 & 39.94 & 45.06 & 34.64 \\
\bf fr  & 41.56 & 21.05 & 39.88 & \bf 41.90 & 18.51 & 27.42 & 40.11 & 44.93 & 33.95 \\
\it ru  & 36.33 & 22.13 & 32.56 & 33.35 & \bf 31.61 & 29.69 & 27.07 & 40.47 & 28.38 \\
\it ja  & 36.28 & 21.17 & 30.36 & 30.60 & 22.26 & \bf 38.65 & 34.26 & 36.83 & 26.81 \\
\bf zh  & 39.91 & 18.48 & 35.53 & 35.68 & 16.44 & 26.36 & \bf 42.04 & 41.94 & 28.93 \\
\bf ar  & 39.60 & 21.49 & 38.29 & 36.87 & 19.58 & 26.15 & 36.76 & \bf 46.23 & 32.70 \\
\bf id  & 40.00 & 21.59 & 38.70 & 37.47 & 19.90 & 26.77 & 35.25 & 42.19 & \bf 38.90 \\
\bottomrule[1.5pt]
\end{tabular}
}
\caption{
Results of language generalization experiments in \indexmain{asym}{asym} setting, with language codes in \textbf{bold} included in the BLOOM pre-training, while the ones in \textit{italic} are not.
Language information refer to Table \ref{tab:lang-brief}.
}
\label{tab:lang-asym}
\end{table}

%% file: tables/scaling.tex
\begin{table}[tb]
\centering
\resizebox{\columnwidth}{!}{
\begin{tabular}{c|ccccc}
\toprule[1.5pt]
Model    & en & zh & ar & id & java \\
\midrule
en-1b1  & 60.31 & 56.46 & 55.55 & 52.08 & 53.92 \\
\midrule
\multicolumn{6}{c}{Scaling model size} \\
\midrule
en-3b~~  & 61.93\textcolor{green}{\small+1.62} & 58.51\textcolor{green}{\small+2.05} & 58.25\textcolor{green}{\small+2.70} & 54.56\textcolor{green}{\small+2.48} & 56.28\textcolor{green}{\small+2.36} \\
en-7b1   & 63.47\textcolor{green}{\small+3.16} & 60.01\textcolor{green}{\small+3.55} & 60.06\textcolor{green}{\small+4.51} & 56.86\textcolor{green}{\small+4.78} & 56.73\textcolor{green}{\small+2.81} \\
\midrule
\multicolumn{6}{c}{Full parameter tuning} \\
\midrule
en-1b1   & 61.55\textcolor{green}{\small+1.24} & 58.98\textcolor{green}{\small+2.42} & 56.53\textcolor{green}{\small+0.98} & 51.68\textcolor{red}{\small-0.4} & 53.53\textcolor{red}{\small-0.39} \\
\bottomrule[1.5pt]
\end{tabular}
}
\caption{
Results of English data trained models of scaling and ablation experiments in \indexmain{all}{all} setting.
}
\label{tab:scaling}
\end{table}

%% file: tables/mteb-en.tex
\begin{table*} 
\centering
\resizebox{\textwidth}{!}{
\begin{tabular}{l|c|cccccccc}
\toprule[1pt]
 & \bf Avg. &  \bf Class. & \bf Clust. & \bf PairClass. & \bf Rerank. & \bf Retr. & \bf STS & \bf Summ.  \\
\bf \#Datasets ($\rightarrow$) & 56 & 12 & 11 & 3 & 4 & 15 & 10 & 1  \\
\midrule
e5-mistral-7b-instruct \citep{wang2024improving} & \bf 66.63   & \bf 78.47 & \bf 50.26 & \bf 88.34 & \bf 60.21 & \bf 56.89 & \bf 84.63 & 31.4 \\
bge-large-en-v1.5 \citep{xiao2023c} & 64.23 & 75.97 & 46.08 & 87.12 & 60.03 & 54.29 & 83.11 & 31.61 \\
SGPT-5.8B-msmarco \citep{muennighoff2022sgpt} & 58.93 & 68.13 & 40.34 & 82    & 56.56 & 50.25 & 78.1  & 31.46 \\
sgpt-bloom-7b1-msmarco \citep{Scao2022BLOOMA1} & 57.59 & 66.19 & 38.93 & 81.9  & 55.65 & 48.22 & 77.74 & \bf 33.6  \\
\midrule
en-all-bloom-1b1    & 58.36   & 69.74 & 40.14 & 83.06 & 53.22 & 45.89 & 80.88 & 30.31 \\
en-all-bloom-3b     & 59.70   & 71.87 & 41.25 & 83.88 & 52.69 & 47.64 & 81.80 & 32.07 \\
en-all-bloom-7b1    & 60.62   & 71.72 & 42.31 & 85.00 & 54.81 & 49.06 & 82.66 & 32.24 \\
\bottomrule[1pt]
\end{tabular}}
\caption{
Results on MTEB English subset. We include the scores of top-performing encoder model, \ie BGE, and deocder-only models from the leaderboard (retrieved on Feb 3th, 2024).
}
\label{tab:mteb-en-results}
\end{table*}

%% file: tables/multicpr.tex
\begin{table*}
\centering
\resizebox{1.0\textwidth}{!}{
\begin{tabular}{@{}c|c|c|cc|cc|cc}
\toprule[1pt]
\multirow{2}{*}{\bf Model} &
  {\multirow{2}{*}{\bf Dataset}} &
  {\multirow{2}{*}{\bf Backbone}} &
  \multicolumn{2}{c|}{\bf E-commerce} &
  \multicolumn{2}{c|}{\bf Entertainment video} &
  \multicolumn{2}{c}{\bf Medical} \\ 
& & &MRR@10 & Recall@1k & MRR@10 & Recall@1k & MRR@10 & Recall@1k \\ \midrule
DPR-1 & In-Domain & BERT & 0.270 & 0.921 & 0.254 & 0.934 & 0.327 & 0.747 \\
DPR-2 & In-Domain & BERT-CT & 0.289 & \bf 0.926 & 0.263 & \bf 0.935 & 0.339  & \bf 0.769 \\
\midrule
text-embedding-ada-002 & General & GPT & 0.183 & 0.825 & 0.159 & 0.786 & 0.245 & 0.593 \\
sgpt-bloom-7b1-msmarco & General & BLOOM & 0.242 & 0.840 & 0.227 & 0.829 & 0.311 & 0.675 \\
\midrule
en-all-bloom-1b1 & General & BLOOM & 0.244 & 0.863 & 0.208 & 0.815 & 0.241  & 0.557 \\
en-all-bloom-3b & General & BLOOM & 0.267 & 0.871 & 0.228 & 0.836 & 0.288  & 0.619 \\
en-all-bloom-7b1 & General & BLOOM & \bf 0.296 & 0.889 & \bf 0.267 & 0.907 & \bf 0.343  & 0.705 \\
\bottomrule[1pt]
\end{tabular}
}
\caption{
Results on Multi-CPR \citep{long2022multicpr}. ``In-Domain'' indicates that the adopted training dataset is from the corresponding domain. ``BERT-CT'' notes that the BERT model is continuing pre-trained with domain corpus.
}\label{tab:multicpr-results}
\end{table*}

%% file: tables/codesearchnet.tex
\begin{table}
\centering
\setlength{\tabcolsep}{4pt}
\resizebox{\columnwidth}{!}{
\tabcolsep=0.105cm
\begin{tabular}{lccccccc}
\toprule[1pt]
& \bf Go & \bf Ruby & \bf Python & \bf Java & \bf JS & \bf PHP & \bf Avg. \\
\midrule
CodeBERT & 69.3 & 70.6 & 84.0 & 86.8 & 74.8 & 70.6 & 76.0 \\
GraphCodeBERT & 84.1 & 73.2 & 87.9 & 75.7 & 71.1 & 72.5 & 77.4 \\
\texttt{cpt-code} S & \textbf{97.7} & \textbf{86.3} & 99.8 & 94.0 & 86.0 & 96.7 & 93.4 \\
\texttt{cpt-code} M & 97.5 & 85.5 & \textbf{99.9} & \textbf{94.4} & \textbf{86.5} & \textbf{97.2} & \textbf{93.5} \\
\texttt{sgpt-bloom-7b1-msmarco} & 76.79 & 69.25 & 95.68 & 77.93 & 70.35 & 73.45 & 77.24 \\
\midrule
en-all-bloom-1b1   & 80.96 & 72.43 & 98.49 & 83.09 & 75.11 & 77.77 & 81.31 \\ 
en-all-bloom-3b    & 81.04 & 76.30 & 98.45 & 84.34 & 77.22 & 79.58 & 82.82 \\ 
en-all-bloom-7b1   & 81.66 & 79.02 & 98.14 & 84.88 & 78.55 & 79.92 & 83.70 \\
\bottomrule[1pt]
\end{tabular}}
\caption{
Results on CodeSearchNet \citep{husain2019codesearchnet}. Scores of CodeBERT \citep{feng-etal-2020-codebert}, GraphCodeBERT \citep{guographcodebert}, and OpenAI API \texttt{cpt-code} are taken from \citet{Neelakantan2022TextAC}.
}
\label{tab:code-search-net}
\end{table}

%% file: tables/sts17-multi.tex
\begin{table}
\centering
\resizebox{\columnwidth}{!}{
\begin{tabular}{l|cccc}
\toprule[1pt]
\bf Model    & \bf ar & \bf en & \bf es & \textit{ko}  \\
\midrule
LASER2                                          & 67.47 & 76.73 & 79.67 & 70.52          \\
LaBSE                                           & 69.07 & 79.45 & 80.83 & 71.32          \\
paraphrase-multilingual-MiniLM-L12-v2           & 79.16 & 86.87 & 85.56 & 77.03          \\
paraphrase-multilingual-mpnet-base-v2           & 79.1  & 86.99 & 85.14 & \bf 83.41       \\
sgpt-bloom-7b1-msmarco                          & 76.42 & 87.07 & 86    & 66.89         \\
multilingual-e5-base                            & 74.52 & 87.83 & 86.74 & 79.95          \\
\midrule
en-all-bloom-1b1   & 81.31 & 89.85 & 86.36 & 61.43   \\
en-all-bloom-3b    & 81.67 & 90.77 & 86.60 & 66.12   \\
en-all-bloom-7b1   & \bf 83.41 & \bf 91.60 & \bf 87.72 & 66.53  \\
\bottomrule[1pt]
\end{tabular}
}\caption{
Spearman correlation between embedding cosine similarity and labels on STS17 multilingual testset.
Language codes in \textit{italic} are not included in the BLOOM pre-training.
Reference results are from MTEB. 
}\label{tab:sts17-multi}
\end{table}

%% file: tables/bucc.tex
\begin{table}
\centering
\resizebox{\columnwidth}{!}{
\begin{tabular}{llllll}
\toprule[1pt]
\bf Model                                 & \bf fr-en & \bf zh-en  &  \textit{de}-\textbf{en} & \textit{ru}-\textbf{en} \\
\midrule
LASER2                                & 98.39 & 97.7   & 99.21 & 97.62 \\
LaBSE                                 & \bf 98.72 & \bf 99.16  & \bf 99.35 & \bf 97.78 \\
multilingual-e5-base                  & 97.59 & 98.3   & 99.13 & 97.20  \\
paraphrase-multilingual-mpnet-base-v2 & 96.89 & 97.56  & 98.59 & 96.44 \\
paraphrase-multilingual-MiniLM-L12-v2 & 94.99 & 95.63  & 97.11 & 95.06 \\
sgpt-bloom-7b1-msmarco                & 97.06 & 97.96  & 54.00    & 45.30  \\
\midrule
en-all-bloom-1b1                      & 97.76 & 97.70 & 38.61 & 23.67 \\
en-all-bloom-3b                       & 98.29 & 98.82 & 71.18 & 66.92 \\
en-all-bloom-7b1                      & 98.52 & 98.77 & 90.11 & 83.74 \\
\bottomrule[1pt]
\end{tabular}
}
\caption{
BUCC F1 scores from MTEB. Languages in \textit{italic} are not included in the BLOOM pre-training.
Baseline results are retrieved from MTEB.
}
\label{tab:bucc}
\end{table}

%% file: tables/main-qwen.tex
\begin{table*}
\resizebox{0.98\textwidth}{!}{
\begin{tabular}{cc|rrrrr|r|rrrrr|r|rrrrr|r}
\toprule[1.5pt]
Setting    &  Eval $\rightarrow$  & \multicolumn{6}{c|}{Asym} & \multicolumn{6}{c|}{Sym} & \multicolumn{6}{c}{All} \\
\cmidrule(lr){3-8} \cmidrule(lr){9-14} \cmidrule(lr){15-20}
Train $\downarrow$ & Lang  & \multicolumn{1}{c}{en} & \multicolumn{1}{c}{zh} & \multicolumn{1}{c}{ar} & \multicolumn{1}{c}{id} & \multicolumn{1}{c}{java} & \multicolumn{1}{|c|}{avg.} & \multicolumn{1}{c}{en} & \multicolumn{1}{c}{zh} & \multicolumn{1}{c}{ar} & \multicolumn{1}{c}{id} & \multicolumn{1}{c}{java} & \multicolumn{1}{|c|}{avg.} & \multicolumn{1}{c}{en} & \multicolumn{1}{c}{zh} & \multicolumn{1}{c}{ar} & \multicolumn{1}{c}{id} & \multicolumn{1}{c}{java} & \multicolumn{1}{|c}{avg.} \\
\midrule \midrule
\multicolumn{20}{c}{BLOOM-1b1} \\
\midrule
\multirow{5}{*}{All}
& en   & 42.97 & 37.96 & 42.85 & 32.09 & 50.70 & 41.31 & 77.65 & 74.95 & 68.26 & 72.06 & 57.14 & 70.01 & \bf 60.31 & 56.46 & 55.55 & 52.08 & 53.92 & \bf 55.66 \\
& zh   & 38.92 & 40.48 & 41.08 & 28.46 & 49.79 & 39.75 & 77.68 & 75.00 & 68.39 & 71.58 & 58.27 & 70.18 & 58.30 & \bf 57.74 & 54.73 & 50.02 & 54.03 & 54.96 \\
& ar   & 38.43 & 36.21 & 45.55 & 32.33 & 49.07 & 40.32 & 77.76 & 75.12 & 69.74 & 73.58 & 57.21 & 70.68 & 58.09 & 55.67 & \bf 57.65 & 52.95 & 53.14 & 55.50 \\
& id   & 39.48 & 34.08 & 41.41 & 38.20 & 48.58 & 40.35 & 77.69 & 74.13 & 68.78 & 75.39 & 56.82 & 70.56 & 58.58 & 54.11 & 55.09 & \bf 56.79 & 52.70 & 55.45 \\
& java & 14.62 & 20.31 & 21.97 & 15.02 & 51.56 & 24.70 & 72.60 & 72.24 & 62.74 & 68.12 & \bf 76.12 & 70.37 & 43.61 & 46.28 & 42.36 & 41.57 & \bf 63.84 & 47.53 \\
\midrule \midrule
\multicolumn{20}{c}{Qwen1.5-0.5B} \\
\midrule
\multirow{5}{*}{All}
& en   &  42.42 & 38.36 & 24.66 & 20.41 & 52.63 & 35.70 & 79.23 & 75.33 & 52.96 & 61.09 & 60.28 & 65.78 & 60.82 & 56.85 & 38.81 & 40.75 & 56.46 & 50.74 \\
& zh   &  40.03 & 41.02 & 24.71 & 17.68 & 53.25 & 35.34 & 78.82 & 75.79 & 52.89 & 60.48 & 61.23 & 65.84 & 59.42 & 58.41 & 38.80 & 39.08 & 57.24 & 50.59 \\
& ar   &  36.32 & 33.34 & 37.64 & 22.85 & 52.25 & 36.48 & 76.85 & 73.43 & 62.32 & 63.02 & 58.77 & 66.88 & 56.59 & 53.38 & 49.98 & 42.94 & 55.51 & 51.68 \\
& id   &  38.22 & 34.97 & 29.67 & 34.54 & 53.81 & 38.24 & 77.32 & 73.68 & 54.96 & 69.85 & 60.44 & 67.25 & 57.77 & 54.32 & 42.32 & 52.20 & 57.12 & 52.75 \\
& java &  18.19 & 24.25 & 2.30 & 5.36 & 50.65 & 20.15 & 71.90 & 70.18 & 44.49 & 54.89 & 75.60 & 63.41 & 45.04 & 47.21 & 23.39 & 30.13 & 63.12 & 41.78 \\
\bottomrule[1.5pt]
\end{tabular}}
\caption{
Main Results of BLOOM-1b1 and Qwen1.5-0.5B. The socre of the \texttt{asym} (or \texttt{sym}) is the macro average of an in-domain test and a out-of-domain test. All tests are listed in \S\ref{sec:exp-design}. The score of the \texttt{all} is the macro average of \texttt{asym} and \texttt{sym}.
}
\label{tab:main-qwen}
\end{table*}

%% file: tables/mteb-en-qwen.tex
\begin{table*} 
\centering
\resizebox{\textwidth}{!}{
\begin{tabular}{l|c|cccccccc}
\toprule[1pt]
 & \bf Avg. &  \bf Class. & \bf Clust. & \bf PairClass. & \bf Rerank. & \bf Retr. & \bf STS & \bf Summ.  \\
\bf \#Datasets ($\rightarrow$) & 56 & 12 & 11 & 3 & 4 & 15 & 10 & 1  \\
\midrule
e5-mistral-7b-instruct \citep{wang2024improving} & \bf 66.63   & \bf 78.47 & \bf 50.26 & \bf 88.34 & \bf 60.21 & \bf 56.89 & \bf 84.63 & 31.4 \\
bge-large-en-v1.5 \citep{xiao2023c} & 64.23 & 75.97 & 46.08 & 87.12 & 60.03 & 54.29 & 83.11 & 31.61 \\
SGPT-5.8B-msmarco \citep{muennighoff2022sgpt} & 58.93 & 68.13 & 40.34 & 82    & 56.56 & 50.25 & 78.1  & 31.46 \\
sgpt-bloom-7b1-msmarco \citep{Scao2022BLOOMA1} & 57.59 & 66.19 & 38.93 & 81.9  & 55.65 & 48.22 & 77.74 & \bf 33.6  \\
\midrule
en-all-bloom-1b1    & 58.36   & 69.74 & 40.14 & 83.06 & 53.22 & 45.89 & 80.88 & 30.31 \\
en-all-bloom-3b     & 59.70   & 71.87 & 41.25 & 83.88 & 52.69 & 47.64 & 81.80 & 32.07 \\
en-all-bloom-7b1    & 60.62   & 71.72 & 42.31 & 85.00 & 54.81 & 49.06 & 82.66 & 32.24 \\
\midrule
en-all-qwen1.5-0.5b    & 58.89   & 71.71 & 39.87 & 83.61 & 53.81 & 46.43 & 80.46 & 31.62 \\
en-all-qwen1.5-1.8b    & 60.73   & 72.83 & 42.91 & 84.75 & 55.19 & 48.79 & 81.66 & 31.31 \\
en-all-qwen1.5-4b      & 62.41   & 74.53 & 44.61 & 85.58 & 55.35 & 51.36 & 82.98 & 31.27 \\
\bottomrule[1pt]
\end{tabular}}
\caption{
Results on MTEB English subset. We include the scores of top-performing encoder model, \ie BGE, and deocder-only models from the leaderboard (retrieved on Feb 3th, 2024).
}
\label{tab:mteb-en-qwen}
\end{table*}

%% file: tables/ablation.tex
\begin{table*}\centering
\resizebox{\textwidth}{!}{
\begin{tabular}{clc|ccccccc}
\toprule
\bf No. & \bf Model Setting & \bf Overall & \bf Class. & \bf Clust. & \bf PairClass. & \bf Rerank. & \bf Retr. & \bf STS & \bf Summ. \\
\midrule
0 & Our-bloom-560m    & 55.80 & 68.04 & 36.89 & 81.05 & 52.60 & 41.19 & 79.93 & 32.06 \\
1 & ~~w/o allnli             & 54.01 & 62.52 & 37.12 & 78.90 & 52.95 & 42.19 & 75.57 & 29.16 \\
2 & ~~w/o msmarco            & 49.14 & 67.74 & 32.84 & 78.81 & 50.02 & 20.78 & 79.98 & 29.84 \\
3 & ~~w/o multiple negatives & 55.70 & 68.19 & 37.30 & 80.60 & 52.87 & 40.63 & 79.63 & 31.49 \\
4 & ~~w/ weightedmean        & 55.37 & 66.60 & 36.42 & 80.26 & 52.98 & 42.14 & 78.89 & 30.58 \\
\midrule
5 & sgpt-bloom-560m        & 53.01 & 62.89 & 36.58 & 76.61 & 52.06 & 39.96 & 74.40 & 30.09 \\
6 & ~~w/ learnable special token + lasttoken pooling & 54.24 & 62.45 & 38.33 & 77.89 & 53.22 & 42.22 & 75.69 & 29.48 \\
\bottomrule
\end{tabular}}
\caption{
Ablation study. MTEB English results of bloom-560m finetuned by different settings.
}\label{tab:ablation}
\end{table*}

%% file: tables/full.tex
\begin{table*}
\resizebox{\textwidth}{!}{
\setlength{\tabcolsep}{3pt} 
\begin{tabular}{cc|r|rrrrrr|rrr|rrr|rrr|rrr}
\toprule[1.5pt]
&  Train $\rightarrow$ & raw & \multicolumn{6}{c}{english}                                    & \multicolumn{3}{|c}{zh}       & \multicolumn{3}{|c}{ar}       & \multicolumn{3}{|c}{id}       & \multicolumn{3}{|c}{java}     \\
 & Eval $\downarrow$ & 1b1   & 1b1-asym & 1b1-sym & 1b1-all & 1b1-all-full & 3b-all & 7b1-all & 1b1-asym & 1b1-sym & 1b1-all & 1b1-asym & 1b1-sym & 1b1-all & 1b1-asym & 1b1-sym & 1b1-all & 1b1-sym & 1b1-asym & 1b1-all \\
\midrule \midrule
\multirow{5}{*}{en}   & mMarco          & 0.01  & 39.79    & 8.8     & 38.49   & 42.72        & 40.49  & 41.98   & 36.21    & 7.94    & 34.99   & 35.86    & 7.45    & 34.24   & 36.34    & 8.7     & 35.83   & 0       & 13.58    & 12.95   \\
                      & Miracl                      & 0     & 47.91    & 3.08    & 47.44   & 48.41        & 48.3   & 50.42   & 43.6     & 2.36    & 42.86   & 43.34    & 4.33    & 42.62   & 43.67    & 6.32    & 43.12   & 0       & 17.15    & 16.29   \\
                      & STSBenchmarkMultilingual    & 12.21 & 79.53    & 85.96   & 85.15   & 85.35        & 86.76  & 87.37   & 78.75    & 86.42   & 84.36   & 78.81    & 84.54   & 84.24   & 79.16    & 85.32   & 84.28   & 23.54   & 73.24    & 73.56   \\
                      & STS17Extend                 & 35.44 & 86.47    & 89.84   & 89.85   & 90.01        & 90.77  & 91.6    & 84.98    & 88.82   & 88.88   & 85.03    & 88.01   & 88.42   & 85.49    & 88.9    & 88.88   & 37.63   & 80.83    & 82.51   \\
                      & MassiveIntentClassification & 28.22 & 67       & 70.92   & 67.8    & 67.38        & 70.18  & 72.01   & 68.24    & 70.06   & 68.75   & 68.31    & 71.01   & 69.18   & 67.7     & 69.72   & 68.8    & 34.75   & 67.5     & 67.16   \\
\midrule
\multirow{5}{*}{zh}   & mMarco          & 0.02  & 27.01    & 8.01    & 26.27   & 30.02        & 28.43  & 29.69   & 31.06    & 6.86    & 30.19   & 27.12    & 7.06    & 26.32   & 25.95    & 5.83    & 25.07   & 0.04    & 12.91    & 13.41   \\
                      & Miracl                      & 0     & 52.84    & 10.92   & 49.66   & 54.14        & 52.75  & 55.69   & 53.03    & 7.65    & 50.77   & 46.41    & 9.31    & 46.1    & 44.55    & 3.56    & 43.09   & 0       & 25.89    & 27.22   \\
                      & STSBenchmarkMultilingual    & 25.41 & 74.62    & 79.59   & 78.89   & 80.68        & 80.82  & 81.49   & 75.83    & 81.65   & 80.72   & 75.47    & 79.66   & 79.13   & 74.4     & 79.26   & 78.05   & 33.03   & 71.09    & 71.52   \\
                      & STS17Extend                 & 38.29 & 81.77    & 85.99   & 86.9    & 87.87        & 88.47  & 88.86   & 83.87    & 87.49   & 87.62   & 82.23    & 85.19   & 86.19   & 80.48    & 84.65   & 84.41   & 41.67   & 79.69    & 79.52   \\
                      & MassiveIntentClassification & 31.75 & 65.8     & 69.67   & 67.01   & 67.49        & 68.22  & 69.5    & 65.51    & 68.72   & 65.82   & 66.78    & 69.59   & 67.59   & 65.95    & 69.29   & 67.03   & 41.5    & 69.25    & 68.95   \\
\midrule
\multirow{5}{*}{ar}   & mMarco          & 0.05  & 22.04    & 4.04    & 21.33   & 24.35        & 23.79  & 25.97   & 22.85    & 5.75    & 22.24   & 27.36    & 5.95    & 26.48   & 23.59    & 7.04    & 22.99   & 0.01    & 8.28     & 9.75    \\
                      & Miracl                      & 0.07  & 65.25    & 5.7     & 64.36   & 63.69        & 68.16  & 70.26   & 61.02    & 7.78    & 59.91   & 65.09    & 11.19   & 64.63   & 60.8     & 13.53   & 59.82   & 0       & 32.6     & 34.19   \\
                      & STSBenchmarkMultilingual    & 29.51 & 69.54    & 75.94   & 75.94   & 79.16        & 79.34  & 81.44   & 72.14    & 78.49   & 77.41   & 73.32    & 79.39   & 79.78   & 73.34    & 77.75   & 77.8    & 20.52   & 66.88    & 67.64   \\
                      & STS17Extend                 & 31.43 & 72.61    & 80.68   & 81.31   & 82.26        & 81.67  & 83.41   & 74.55    & 80.53   & 80.9    & 76.7     & 83.38   & 84.17   & 76.74    & 80.27   & 81.76   & 16.35   & 67.29    & 66.26   \\
                      & MassiveIntentClassification & 19.08 & 56.46    & 59.44   & 57.88   & 57.38        & 60.53  & 61.57   & 57.29    & 58.02   & 57.62   & 56.45    & 59.4    & 57.51   & 56.43    & 58.41   & 57.77   & 28.1    & 58.6     & 58.53   \\
\midrule
\multirow{5}{*}{id}   & mMarco          & 0.01  & 20.04    & 4.89    & 21.41   & 21.92        & 26.16  & 29.26   & 19.32    & 4.97    & 18.97   & 24.86    & 5.06    & 24.16   & 33.03    & 6.29    & 32.03   & 0.01    & 6.92     & 6.67    \\
                      & Miracl                      & 0     & 42.82    & 6.71    & 42.77   & 40.42        & 44.2   & 45.85   & 38.54    & 8.78    & 37.95   & 40.54    & 9.69    & 40.49   & 44.77    & 10.47   & 44.36   & 0.03    & 20.13    & 23.38   \\
                      & STSBenchmarkMultilingual    & 24.91 & 72.11    & 79.58   & 78.36   & 80.72        & 81.03  & 83.2    & 72.73    & 81.06   & 78.75   & 73.1     & 80.63   & 79.78   & 76.89    & 83.13   & 82.91   & 24.12   & 69.54    & 69.4    \\
                      & STS17Extend                 & 47.12 & 80.32    & 86.55   & 86.25   & 88.51        & 87.87  & 89.63   & 79.19    & 86      & 84.31   & 81.1     & 86.77   & 87.28   & 83.53    & 87.98   & 88.98   & 44.45   & 77.11    & 76.83   \\
                      & MassiveIntentClassification & 22.7  & 60.81    & 64.77   & 61.82   & 59.77        & 63.43  & 65.91   & 61.18    & 63.67   & 61.62   & 62.6     & 66.09   & 63.63   & 63.54    & 66.79   & 64.83   & 32.74   & 63.42    & 63.13   \\
\midrule
\multirow{5}{*}{java} & CodeSearchNet               & 1.00   & 82.45    & 73.27   & 83.09   & 82.84        & 84.33  & 84.87   & 82.77    & 75.17   & 82.64   & 82.4     & 73.81   & 81.66   & 81.1     & 62.46   & 81.41   & 3.14    & 88.53    & 88.47   \\
                      & xCodeEvalRetrievalNlCode    & 0     & 12.74    & 11.4    & 18.31   & 15.94        & 20.06  & 20.43   & 15.72    & 11.08   & 16.94   & 17.78    & 11.91   & 16.48   & 15.7     & 9.84    & 15.76   & 0       & 17.47    & 14.64   \\
                      & BigCloneBench               & 19.14 & 48.05    & 43.83   & 45.96   & 48.67        & 50.76  & 50.18   & 47.53    & 44.71   & 47.77   & 44.19    & 43.97   & 45.63   & 44.79    & 42.4    & 45.42   & 94.61   & 46.81    & 95.48   \\
                      & GoogleCodeJam               & 61.79 & 67.43    & 68.28   & 68.33   & 66.67        & 69.98  & 71.45   & 69.55    & 69.17   & 68.78   & 69.67    & 67.57   & 68.8    & 70.95    & 66.79   & 68.22   & 52.07   & 62.72    & 56.77  \\
\bottomrule[1.5pt]
\end{tabular}}
\caption{
Detailed results of Table \ref{tab:main} on our compiled universal embedding benchmark. \texttt{raw-1b1} is un-finetuned BLOOM 1b1 model tested with <EOS> embeddings.
}
\label{tab:main-full}
\end{table*}

%% file: tables/full-qwen.tex
\begin{table}
\centering
\resizebox{\columnwidth}{!}{
\begin{tabular}{ll|rrrrr}
\toprule[1pt]
&  & \multicolumn{1}{l}{en-all} & \multicolumn{1}{l}{zh-all} & \multicolumn{1}{l}{ar-all} & \multicolumn{1}{l}{id-all} & \multicolumn{1}{l}{java-all} \\ \midrule
\multirow{5}{*}{en}   & mMarcoMultilingual          & 38.56                                        & 36.06                                        & 33.01                                       & 34.30                                           & 15.65                                     \\
                      & Miracl                      & 46.28                                        & 44.00                                        & 39.63                                       & 42.14                                           & 20.73                                     \\
                      & STSBenchmarkMultilingual    & 84.64                                        & 84.30                                        & 79.28                                       & 81.22                                           & 71.93                                     \\
                      & STS17Extend                 & 90.80                                        & 90.20                                        & 88.08                                       & 88.29                                           & 77.70                                     \\
                      & MassiveIntentClassification & 70.73                                        & 70.39                                        & 70.02                                       & 69.89                                           & 68.97                                     \\
\midrule
\multirow{5}{*}{zh}   & mMarcoMultilingual          & 26.14                                        & 29.51                                        & 23.19                                       & 23.79                                           & 13.69                                     \\
                      & Miracl                      & 50.58                                        & 52.53                                        & 43.48                                       & 46.15                                           & 34.80                                     \\
                      & STSBenchmarkMultilingual    & 77.57                                        & 79.79                                        & 72.53                                       & 74.51                                           & 68.07                                     \\
                      & STS17Extend                 & 88.42                                        & 89.15                                        & 84.89                                       & 85.27                                           & 76.85                                     \\
                      & MassiveIntentClassification & 67.67                                        & 67.11                                        & 68.15                                       & 67.47                                           & 67.90                                     \\
\midrule
\multirow{5}{*}{ar}   & mMarcoMultilingual          & 12.40                                        & 12.79                                        & 21.52                                       & 15.84                                           & 1.80                                      \\
                      & Miracl                      & 36.92                                        & 36.63                                        & 53.76                                       & 43.51                                           & 2.79                                      \\
                      & STSBenchmarkMultilingual    & 62.27                                        & 62.47                                        & 73.10                                       & 64.17                                           & 54.03                                     \\
                      & STS17Extend                 & 59.46                                        & 58.79                                        & 77.54                                       & 64.59                                           & 43.90                                     \\
                      & MassiveIntentClassification & 45.06                                        & 45.14                                        & 49.32                                       & 45.54                                           & 40.02                                     \\
\midrule
\multirow{5}{*}{id}   & mMarcoMultilingual          & 14.54                                        & 13.36                                        & 16.57                                       & 27.53                                           & 3.17                                      \\
                      & Miracl                      & 26.28                                        & 22.01                                        & 29.13                                       & 41.55                                           & 7.55                                      \\
                      & STSBenchmarkMultilingual    & 65.61                                        & 63.97                                        & 66.63                                       & 77.18                                           & 54.28                                     \\
                      & STS17Extend                 & 71.77                                        & 72.19                                        & 76.81                                       & 86.16                                           & 65.59                                     \\
                      & MassiveIntentClassification & 53.48                                        & 52.87                                        & 54.32                                       & 58.03                                           & 49.85                                     \\
\midrule
\multirow{4}{*}{java} & CodeSearchNet               & 83.95                                        & 83.00                                        & 82.47                                       & 83.00                                           & 88.25                                     \\
                      & xCodeEvalRetrievalNlCode    & 21.31                                        & 23.51                                        & 22.03                                       & 24.62                                           & 13.04                                     \\
                      & BigCloneBench               & 48.56                                        & 50.68                                        & 45.95                                       & 48.18                                           & 96.85                                     \\
                      & GoogleCodeJam               & 72.00                                        & 71.78                                        & 71.59                                       & 72.69                                           & 54.35                                    \\
\bottomrule[1pt]
\end{tabular}}
\caption{
Detailed results of Qwen1.5-0.5B of Table \ref{tab:main-qwen}.
}
\end{table}

%% file: acl_latex.bbl
\begin{thebibliography}{73}
\providecommand{\natexlab}[1]{#1}

\bibitem[{Artetxe and Schwenk(2019)}]{artetxe-schwenk-2019-massively}
Mikel Artetxe and Holger Schwenk. 2019.
\newblock \href {https://doi.org/10.1162/tacl\_a\_00288} {Massively multilingual sentence embeddings for zero-shot cross-lingual transfer and beyond}.
\newblock \emph{Transactions of the Association for Computational Linguistics}, 7:597--610.

\bibitem[{Badola et~al.(2023)Badola, Dave, and Talukdar}]{badola-etal-2023-parameter}
Kartikeya Badola, Shachi Dave, and Partha Talukdar. 2023.
\newblock \href {https://doi.org/10.18653/v1/2023.findings-acl.619} {Parameter-efficient finetuning for robust continual multilingual learning}.
\newblock In \emph{Findings of the Association for Computational Linguistics: ACL 2023}, pages 9763--9780, Toronto, Canada. Association for Computational Linguistics.

\bibitem[{Ben~Zaken et~al.(2022)Ben~Zaken, Goldberg, and Ravfogel}]{ben-zaken-etal-2022-bitfit}
Elad Ben~Zaken, Yoav Goldberg, and Shauli Ravfogel. 2022.
\newblock \href {https://doi.org/10.18653/v1/2022.acl-short.1} {{B}it{F}it: Simple parameter-efficient fine-tuning for transformer-based masked language-models}.
\newblock In \emph{Proc. of the ACL}, pages 1--9, Dublin, Ireland.

\bibitem[{Bonifacio et~al.(2021)Bonifacio, Jeronymo, Abonizio, Campiotti, Fadaee, Lotufo, and Nogueira}]{bonifacio2021mmarco}
Luiz Bonifacio, Vitor Jeronymo, Hugo~Queiroz Abonizio, Israel Campiotti, Marzieh Fadaee, Roberto Lotufo, and Rodrigo Nogueira. 2021.
\newblock mmarco: A multilingual version of the ms marco passage ranking dataset.
\newblock \emph{arXiv preprint arXiv:2108.13897}.

\bibitem[{Bowman et~al.(2015)Bowman, Angeli, Potts, and Manning}]{bowman-etal-2015-large}
Samuel~R. Bowman, Gabor Angeli, Christopher Potts, and Christopher~D. Manning. 2015.
\newblock \href {https://doi.org/10.18653/v1/D15-1075} {A large annotated corpus for learning natural language inference}.
\newblock In \emph{Proc. of the EMNLP}, pages 632--642, Lisbon, Portugal.

\bibitem[{Cer et~al.(2017)Cer, Diab, Agirre, Lopez-Gazpio, and Specia}]{cer-etal-2017-semeval}
Daniel Cer, Mona Diab, Eneko Agirre, I{\~n}igo Lopez-Gazpio, and Lucia Specia. 2017.
\newblock \href {https://doi.org/10.18653/v1/S17-2001} {{S}em{E}val-2017 task 1: Semantic textual similarity multilingual and crosslingual focused evaluation}.
\newblock In \emph{Proceedings of the {S}em{E}val}, pages 1--14, Vancouver, Canada.

\bibitem[{Cer et~al.(2018)Cer, Yang, Kong, Hua, Limtiaco, St.~John, Constant, Guajardo-Cespedes, Yuan, Tar et~al.}]{cer-etal-2018-universal}
Daniel Cer, Yinfei Yang, Sheng-yi Kong, Nan Hua, Nicole Limtiaco, Rhomni St.~John, Noah Constant, Mario Guajardo-Cespedes, Steve Yuan, Chris Tar, and 1 others. 2018.
\newblock \href {https://doi.org/10.18653/v1/D18-2029} {Universal sentence encoder for {E}nglish}.
\newblock In \emph{Proc. of the EMNLP: System Demonstrations}, pages 169--174, Brussels, Belgium.

\bibitem[{Chen et~al.(2024)Chen, Xiao, Zhang, Luo, Lian, and Liu}]{chen2024bgem3}
Jianlv Chen, Shitao Xiao, Peitian Zhang, Kun Luo, Defu Lian, and Zheng Liu. 2024.
\newblock \href {https://arxiv.org/pdf/2402.03216.pdf} {Bge m3-embedding: Multi-lingual, multi-functionality, multi-granularity text embeddings through self-knowledge distillation}.

\bibitem[{Chen et~al.(2020)Chen, Kornblith, Norouzi, and Hinton}]{pmlr-v119-chen20j}
Ting Chen, Simon Kornblith, Mohammad Norouzi, and Geoffrey Hinton. 2020.
\newblock \href {https://proceedings.mlr.press/v119/chen20j.html} {A simple framework for contrastive learning of visual representations}.
\newblock In \emph{Proc. of the ICML}, volume 119, pages 1597--1607.

\bibitem[{Cheng et~al.(2023)Cheng, Yang, Sun, Li, and Qiu}]{cheng2023improving}
Qinyuan Cheng, Xiaogui Yang, Tianxiang Sun, Linyang Li, and Xipeng Qiu. 2023.
\newblock \href {https://arxiv.org/abs/2305.01918} {Improving contrastive learning of sentence embeddings from ai feedback}.
\newblock In \emph{Findings of the ACL}.

\bibitem[{Chidambaram et~al.(2019)Chidambaram, Yang, Cer, Yuan, Sung, Strope, and Kurzweil}]{chidambaram-etal-2019-learning}
Muthu Chidambaram, Yinfei Yang, Daniel Cer, Steve Yuan, Yunhsuan Sung, Brian Strope, and Ray Kurzweil. 2019.
\newblock \href {https://doi.org/10.18653/v1/W19-4330} {Learning cross-lingual sentence representations via a multi-task dual-encoder model}.
\newblock In \emph{Proc. of the 4th Workshop on Representation Learning for NLP (RepL4NLP-2019)}, pages 250--259, Florence, Italy.

\bibitem[{Conneau et~al.(2017)Conneau, Kiela, Schwenk, Barrault, and Bordes}]{conneau-etal-2017-supervised}
Alexis Conneau, Douwe Kiela, Holger Schwenk, Lo{\"\i}c Barrault, and Antoine Bordes. 2017.
\newblock \href {https://doi.org/10.18653/v1/D17-1070} {Supervised learning of universal sentence representations from natural language inference data}.
\newblock In \emph{Proc. of the EMNLP}, pages 670--680, Copenhagen, Denmark.

\bibitem[{Dao(2024)}]{dao-2024-flash}
Tri Dao. 2024.
\newblock Flash{A}ttention-2: Faster attention with better parallelism and work partitioning.
\newblock In \emph{International Conference on Learning Representations}.

\bibitem[{Devlin et~al.(2019)Devlin, Chang, Lee, and Toutanova}]{devlin-etal-2019-bert}
Jacob Devlin, Ming-Wei Chang, Kenton Lee, and Kristina Toutanova. 2019.
\newblock \href {https://doi.org/10.18653/v1/N19-1423} {{BERT}: Pre-training of deep bidirectional transformers for language understanding}.
\newblock In \emph{Proc. of the NAACL-HLT}, pages 4171--4186, Minneapolis, Minnesota. Association for Computational Linguistics.

\bibitem[{Dong et~al.(2021)Dong, Cordonnier, and Loukas}]{pmlr-v139-dong21a}
Yihe Dong, Jean-Baptiste Cordonnier, and Andreas Loukas. 2021.
\newblock \href {https://proceedings.mlr.press/v139/dong21a.html} {Attention is not all you need: pure attention loses rank doubly exponentially with depth}.
\newblock In \emph{Proc. of the ICML}, volume 139, pages 2793--2803.

\bibitem[{Feng et~al.(2022)Feng, Yang, Cer, Arivazhagan, and Wang}]{feng-etal-2022-language}
Fangxiaoyu Feng, Yinfei Yang, Daniel Cer, Naveen Arivazhagan, and Wei Wang. 2022.
\newblock \href {https://doi.org/10.18653/v1/2022.acl-long.62} {Language-agnostic {BERT} sentence embedding}.
\newblock In \emph{Proc. of the ACL}, pages 878--891, Dublin, Ireland.

\bibitem[{Feng et~al.(2020)Feng, Guo, Tang, Duan, Feng, Gong, Shou, Qin, Liu, Jiang, and Zhou}]{feng-etal-2020-codebert}
Zhangyin Feng, Daya Guo, Duyu Tang, Nan Duan, Xiaocheng Feng, Ming Gong, Linjun Shou, Bing Qin, Ting Liu, Daxin Jiang, and Ming Zhou. 2020.
\newblock \href {https://doi.org/10.18653/v1/2020.findings-emnlp.139} {{C}ode{BERT}: A pre-trained model for programming and natural languages}.
\newblock In \emph{Findings of the EMNLP}, pages 1536--1547, Online.

\bibitem[{FitzGerald et~al.(2022)FitzGerald, Hench, Peris, Mackie, Rottmann, Sanchez, Nash, Urbach, Kakarala, Singh et~al.}]{fitzgerald2022massive}
Jack FitzGerald, Christopher Hench, Charith Peris, Scott Mackie, Kay Rottmann, Ana Sanchez, Aaron Nash, Liam Urbach, Vishesh Kakarala, Richa Singh, and 1 others. 2022.
\newblock \href {https://arxiv.org/abs/2204.08582} {Massive: A 1m-example multilingual natural language understanding dataset with 51 typologically-diverse languages}.
\newblock \emph{arXiv preprint arXiv:2204.08582}.

\bibitem[{Gao and Callan(2021)}]{gao-callan-2021-condenser}
Luyu Gao and Jamie Callan. 2021.
\newblock \href {https://doi.org/10.18653/v1/2021.emnlp-main.75} {Condenser: a pre-training architecture for dense retrieval}.
\newblock In \emph{Proc. of the EMNLP}, pages 981--993, Online and Punta Cana, Dominican Republic.

\bibitem[{Gao et~al.(2021{\natexlab{a}})Gao, Zhang, Han, and Callan}]{gao-etal-2021-scaling}
Luyu Gao, Yunyi Zhang, Jiawei Han, and Jamie Callan. 2021{\natexlab{a}}.
\newblock \href {https://doi.org/10.18653/v1/2021.repl4nlp-1.31} {Scaling deep contrastive learning batch size under memory limited setup}.
\newblock In \emph{Proc. of the 6th Workshop on Representation Learning for NLP (RepL4NLP-2021)}, pages 316--321, Online.

\bibitem[{Gao et~al.(2021{\natexlab{b}})Gao, Yao, and Chen}]{gao-etal-2021-simcse}
Tianyu Gao, Xingcheng Yao, and Danqi Chen. 2021{\natexlab{b}}.
\newblock \href {https://doi.org/10.18653/v1/2021.emnlp-main.552} {{S}im{CSE}: Simple contrastive learning of sentence embeddings}.
\newblock In \emph{Proc. of the EMNLP}, pages 6894--6910, Online and Punta Cana, Dominican Republic.

\bibitem[{Giorgi et~al.(2021)Giorgi, Nitski, Wang, and Bader}]{giorgi-etal-2021-declutr}
John Giorgi, Osvald Nitski, Bo~Wang, and Gary Bader. 2021.
\newblock \href {https://doi.org/10.18653/v1/2021.acl-long.72} {{D}e{CLUTR}: Deep contrastive learning for unsupervised textual representations}.
\newblock In \emph{Proc. of the ACL-IJCNLP}, pages 879--895, Online.

\bibitem[{Goswami et~al.(2021)Goswami, Dutta, Assem, Fransen, and McCrae}]{goswami-etal-2021-cross}
Koustava Goswami, Sourav Dutta, Haytham Assem, Theodorus Fransen, and John~P. McCrae. 2021.
\newblock \href {https://doi.org/10.18653/v1/2021.emnlp-main.716} {Cross-lingual sentence embedding using multi-task learning}.
\newblock In \emph{Proc. of EMNLP}, pages 9099--9113, Online and Punta Cana, Dominican Republic.

\bibitem[{Guo et~al.(2021)Guo, Ren, Lu, Feng, Tang, Shujie, Zhou, Duan, Svyatkovskiy, Fu et~al.}]{guographcodebert}
Daya Guo, Shuo Ren, Shuai Lu, Zhangyin Feng, Duyu Tang, LIU Shujie, Long Zhou, Nan Duan, Alexey Svyatkovskiy, Shengyu Fu, and 1 others. 2021.
\newblock \href {https://openreview.net/forum?id=jLoC4ez43PZ} {Graphcodebert: Pre-training code representations with data flow}.
\newblock In \emph{International Conference on Learning Representations}.

\bibitem[{Hill et~al.(2016)Hill, Cho, and Korhonen}]{hill-etal-2016-learning}
Felix Hill, Kyunghyun Cho, and Anna Korhonen. 2016.
\newblock \href {https://doi.org/10.18653/v1/N16-1162} {Learning distributed representations of sentences from unlabelled data}.
\newblock In \emph{Proc. of the NAACL-HLT}, pages 1367--1377, San Diego, California.

\bibitem[{Hu et~al.(2021)Hu, Wallis, Allen-Zhu, Li, Wang, Wang, Chen et~al.}]{hu2021lora}
Edward~J Hu, Phillip Wallis, Zeyuan Allen-Zhu, Yuanzhi Li, Shean Wang, Lu~Wang, Weizhu Chen, and 1 others. 2021.
\newblock Lora: Low-rank adaptation of large language models.
\newblock In \emph{International Conference on Learning Representations}.

\bibitem[{Husain et~al.(2019)Husain, Wu, Gazit, Allamanis, and Brockschmidt}]{husain2019codesearchnet}
Hamel Husain, Ho-Hsiang Wu, Tiferet Gazit, Miltiadis Allamanis, and Marc Brockschmidt. 2019.
\newblock \href {https://arxiv.org/abs/1909.09436} {Codesearchnet challenge: Evaluating the state of semantic code search}.
\newblock \emph{arXiv preprint arXiv:1909.09436}.

\bibitem[{Jiang et~al.(2023)Jiang, Sablayrolles, Mensch, Bamford, Chaplot, Casas, Bressand, Lengyel, Lample, Saulnier et~al.}]{jiang2023mistral}
Albert~Q Jiang, Alexandre Sablayrolles, Arthur Mensch, Chris Bamford, Devendra~Singh Chaplot, Diego de~las Casas, Florian Bressand, Gianna Lengyel, Guillaume Lample, Lucile Saulnier, and 1 others. 2023.
\newblock Mistral 7b.
\newblock \emph{arXiv preprint arXiv:2310.06825}.

\bibitem[{Karpukhin et~al.(2020)Karpukhin, Oguz, Min, Lewis, Wu, Edunov, Chen, and Yih}]{karpukhin-etal-2020-dense}
Vladimir Karpukhin, Barlas Oguz, Sewon Min, Patrick Lewis, Ledell Wu, Sergey Edunov, Danqi Chen, and Wen-tau Yih. 2020.
\newblock \href {https://doi.org/10.18653/v1/2020.emnlp-main.550} {Dense passage retrieval for open-domain question answering}.
\newblock In \emph{Proc. of the EMNLP}, pages 6769--6781, Online.

\bibitem[{Khan et~al.(2023)Khan, Bari, Do, Wang, Parvez, and Joty}]{khan2023xcodeeval}
Mohammad Abdullah~Matin Khan, M~Saiful Bari, Xuan~Long Do, Weishi Wang, Md~Rizwan Parvez, and Shafiq Joty. 2023.
\newblock xcodeeval: A large scale multilingual multitask benchmark for code understanding, generation, translation and retrieval.
\newblock \emph{arXiv preprint arXiv:2303.03004}.

\bibitem[{Kim et~al.(2021)Kim, Yoo, and Lee}]{kim-etal-2021-self}
Taeuk Kim, Kang~Min Yoo, and Sang-goo Lee. 2021.
\newblock \href {https://doi.org/10.18653/v1/2021.acl-long.197} {Self-guided contrastive learning for {BERT} sentence representations}.
\newblock In \emph{Proc. of the ACL-IJCNLP}, pages 2528--2540, Online.

\bibitem[{Lauren{\c{c}}on et~al.(2022)Lauren{\c{c}}on, Saulnier, Wang, Akiki, Villanova~del Moral, Le~Scao, Von~Werra, Mou, Gonz{\'a}lez~Ponferrada, Nguyen et~al.}]{laurenccon2022bigscience}
Hugo Lauren{\c{c}}on, Lucile Saulnier, Thomas Wang, Christopher Akiki, Albert Villanova~del Moral, Teven Le~Scao, Leandro Von~Werra, Chenghao Mou, Eduardo Gonz{\'a}lez~Ponferrada, Huu Nguyen, and 1 others. 2022.
\newblock The bigscience roots corpus: A 1.6 tb composite multilingual dataset.
\newblock \emph{Advances in Neural Information Processing Systems}, 35:31809--31826.

\bibitem[{Li et~al.(2022)Li, Zhao, and Moens}]{li2022brief}
Ruiqi Li, Xiang Zhao, and Marie-Francine Moens. 2022.
\newblock \href {https://dl.acm.org/doi/abs/10.1145/3482853} {A brief overview of universal sentence representation methods: A linguistic view}.
\newblock \emph{ACM Computing Surveys (CSUR)}, 55(3):1--42.

\bibitem[{Li and Li(2023)}]{li2023angle}
Xianming Li and Jing Li. 2023.
\newblock Angle-optimized text embeddings.
\newblock \emph{arXiv preprint arXiv:2309.12871}.

\bibitem[{Li et~al.(2023)Li, Zhang, Zhang, Long, Xie, and Zhang}]{li2023towards}
Zehan Li, Xin Zhang, Yanzhao Zhang, Dingkun Long, Pengjun Xie, and Meishan Zhang. 2023.
\newblock Towards general text embeddings with multi-stage contrastive learning.
\newblock \emph{arXiv preprint arXiv:2308.03281}.

\bibitem[{Lin et~al.(2017)Lin, Feng, Santos, Yu, Xiang, Zhou, and Bengio}]{lin2017structured}
Zhouhan Lin, Minwei Feng, Cicero Nogueira~dos Santos, Mo~Yu, Bing Xiang, Bowen Zhou, and Yoshua Bengio. 2017.
\newblock \href {https://arxiv.org/abs/1703.03130} {A structured self-attentive sentence embedding}.
\newblock \emph{arXiv preprint arXiv:1703.03130}.

\bibitem[{Liu et~al.(2021)Liu, Vuli{\'c}, Korhonen, and Collier}]{liu-etal-2021-fast}
Fangyu Liu, Ivan Vuli{\'c}, Anna Korhonen, and Nigel Collier. 2021.
\newblock \href {https://doi.org/10.18653/v1/2021.emnlp-main.109} {Fast, effective, and self-supervised: Transforming masked language models into universal lexical and sentence encoders}.
\newblock In \emph{Proc. of the EMNLP}, pages 1442--1459, Online and Punta Cana, Dominican Republic.

\bibitem[{Long et~al.(2022)Long, Gao, Zou, Xu, Xie, Guo, Xu, Jiang, Xing, and Yang}]{long2022multicpr}
Dingkun Long, Qiong Gao, Kuan Zou, Guangwei Xu, Pengjun Xie, Ruijie Guo, Jian Xu, Guanjun Jiang, Luxi Xing, and Ping Yang. 2022.
\newblock \href {https://doi.org/10.1145/3477495.3531736} {Multi-cpr: {A} multi domain chinese dataset for passage retrieval}.
\newblock In \emph{Proc. of the SIGIR}, pages 3046--3056. {ACM}.

\bibitem[{Mao et~al.(2022)Mao, Liang, Duan, Wang, Wang, Chen, and Gao}]{mao2022less}
Yuren Mao, Yaobo Liang, Nan Duan, Haobo Wang, Kai Wang, Lu~Chen, and Yunjun Gao. 2022.
\newblock \href {http://papers.nips.cc/paper\_files/paper/2022/hash/5f9f9e4da57a94547491a39dc18f1696-Abstract-Conference.html} {Less-forgetting multi-lingual fine-tuning}.
\newblock In \emph{NeurIPS}.

\bibitem[{McInnes et~al.(2018)McInnes, Healy, Saul, and Gro{\ss}berger}]{mcinnes2018umap}
Leland McInnes, John Healy, Nathaniel Saul, and Lukas Gro{\ss}berger. 2018.
\newblock Umap: Uniform manifold approximation and projection.
\newblock \emph{Journal of Open Source Software}, 3(29):861.

\bibitem[{Muennighoff(2022)}]{muennighoff2022sgpt}
Niklas Muennighoff. 2022.
\newblock \href {https://arxiv.org/abs/2202.08904} {{SGPT}: {GPT} sentence embeddings for semantic search}.
\newblock \emph{arXiv preprint arXiv:2202.08904}.

\bibitem[{Muennighoff et~al.(2023)Muennighoff, Tazi, Magne, and Reimers}]{muennighoff-etal-2023-mteb}
Niklas Muennighoff, Nouamane Tazi, Loic Magne, and Nils Reimers. 2023.
\newblock \href {https://doi.org/10.18653/v1/2023.eacl-main.148} {{MTEB}: Massive text embedding benchmark}.
\newblock In \emph{Proc. of the EACL}, pages 2014--2037, Dubrovnik, Croatia.

\bibitem[{Neelakantan et~al.(2022)Neelakantan, Xu, Puri, Radford, Han, Tworek, Yuan, Tezak, Kim, Hallacy et~al.}]{Neelakantan2022TextAC}
Arvind Neelakantan, Tao Xu, Raul Puri, Alec Radford, Jesse~Michael Han, Jerry Tworek, Qiming Yuan, Nikolas~A. Tezak, Jong~Wook Kim, Chris Hallacy, and 1 others. 2022.
\newblock \href {https://arxiv.org/abs/2201.10005} {Text and code embeddings by contrastive pre-training}.
\newblock \emph{ArXiv}, abs/2201.10005.

\bibitem[{Nguyen et~al.(2016)Nguyen, Rosenberg, Song, Gao, Tiwary, Majumder, and Deng}]{nguyen2016msmarco}
Tri Nguyen, Mir Rosenberg, Xia Song, Jianfeng Gao, Saurabh Tiwary, Rangan Majumder, and Li~Deng. 2016.
\newblock \href {https://ceur-ws.org/Vol-1773/CoCoNIPS\_2016\_paper9.pdf} {{MS} {MARCO:} {A} human generated machine reading comprehension dataset}.
\newblock In \emph{Proceedings of the Workshop on Cognitive Computation: Integrating neural and symbolic approaches}, volume 1773.

\bibitem[{Ni et~al.(2022)Ni, Hernandez~Abrego, Constant, Ma, Hall, Cer, and Yang}]{ni-etal-2022-sentence}
Jianmo Ni, Gustavo Hernandez~Abrego, Noah Constant, Ji~Ma, Keith Hall, Daniel Cer, and Yinfei Yang. 2022.
\newblock \href {https://doi.org/10.18653/v1/2022.findings-acl.146} {Sentence-t5: Scalable sentence encoders from pre-trained text-to-text models}.
\newblock In \emph{Findings of the ACL}, pages 1864--1874, Dublin, Ireland.

\bibitem[{Pagliardini et~al.(2018)Pagliardini, Gupta, and Jaggi}]{pagliardini-etal-2018-unsupervised}
Matteo Pagliardini, Prakhar Gupta, and Martin Jaggi. 2018.
\newblock \href {https://doi.org/10.18653/v1/N18-1049} {Unsupervised learning of sentence embeddings using compositional n-gram features}.
\newblock In \emph{Proc. of the NAACL-HLT}, pages 528--540, New Orleans, Louisiana.

\bibitem[{Peng et~al.(2023)Peng, Galley, He, Cheng, Xie, Hu, Huang, Liden, Yu, Chen et~al.}]{peng2023check}
Baolin Peng, Michel Galley, Pengcheng He, Hao Cheng, Yujia Xie, Yu~Hu, Qiuyuan Huang, Lars Liden, Zhou Yu, Weizhu Chen, and 1 others. 2023.
\newblock \href {https://arxiv.org/abs/2302.12813} {Check your facts and try again: Improving large language models with external knowledge and automated feedback}.
\newblock \emph{arXiv preprint arXiv:2302.12813}.

\bibitem[{Radford et~al.(2018)Radford, Narasimhan, Salimans, Sutskever et~al.}]{radford2018improving}
Alec Radford, Karthik Narasimhan, Tim Salimans, Ilya Sutskever, and 1 others. 2018.
\newblock Improving language understanding by generative pre-training.

\bibitem[{Raffel et~al.(2020)Raffel, Shazeer, Roberts, Lee, Narang, Matena, Zhou, Li, and Liu}]{JMLR:v21:20-074}
Colin Raffel, Noam Shazeer, Adam Roberts, Katherine Lee, Sharan Narang, Michael Matena, Yanqi Zhou, Wei Li, and Peter~J. Liu. 2020.
\newblock \href {http://jmlr.org/papers/v21/20-074.html} {Exploring the limits of transfer learning with a unified text-to-text transformer}.
\newblock \emph{Journal of Machine Learning Research}, 21(140):1--67.

\bibitem[{Reimers and Gurevych(2019)}]{reimers-gurevych-2019-sentence}
Nils Reimers and Iryna Gurevych. 2019.
\newblock \href {https://doi.org/10.18653/v1/D19-1410} {Sentence-{BERT}: Sentence embeddings using {S}iamese {BERT}-networks}.
\newblock In \emph{Proc. of the EMNLP-IJCNLP}, pages 3982--3992, Hong Kong, China.

\bibitem[{Reimers and Gurevych(2020)}]{reimers-gurevych-2020-making}
Nils Reimers and Iryna Gurevych. 2020.
\newblock \href {https://doi.org/10.18653/v1/2020.emnlp-main.365} {Making monolingual sentence embeddings multilingual using knowledge distillation}.
\newblock In \emph{Proc. of the EMNLP}, pages 4512--4525, Online.

\bibitem[{Ruder et~al.(2019)Ruder, Vuli\'{c}, and S\o{}gaard}]{ruder-et-al-2019-asurvey}
Sebastian Ruder, Ivan Vuli\'{c}, and Anders S\o{}gaard. 2019.
\newblock \href {https://doi.org/10.1613/jair.1.11640} {A survey of cross-lingual word embedding models}.
\newblock \emph{J. Artif. Int. Res.}, 65(1):569–630.

\bibitem[{Scao et~al.(2022)Scao, Fan, Akiki, Pavlick, Ili'c, Hesslow, Castagn'e, Luccioni, Yvon, Gall{\'e} et~al.}]{Scao2022BLOOMA1}
Teven~Le Scao, Angela Fan, Christopher Akiki, Elizabeth-Jane Pavlick, Suzana Ili'c, Daniel Hesslow, Roman Castagn'e, Alexandra~Sasha Luccioni, Franccois Yvon, Matthias Gall{\'e}, and 1 others. 2022.
\newblock \href {https://arxiv.org/abs/2211.05100} {Bloom: A 176b-parameter open-access multilingual language model}.
\newblock \emph{ArXiv}, abs/2211.05100.

\bibitem[{Sedykh et~al.(2023)Sedykh, Abulkhanov, Sorokin, Nikolenko, and Malykh}]{sedykh2023searching}
Ivan Sedykh, Dmitry Abulkhanov, Nikita Sorokin, Sergey Nikolenko, and Valentin Malykh. 2023.
\newblock Searching by code: a new searchbysnippet dataset and snipper retrieval model for searching by code snippets.
\newblock \emph{arXiv preprint arXiv:2305.11625}.

\bibitem[{Song et~al.(2022)Song, Wu, Washington, Sadler, Chao, and Su}]{song2022llm}
Chan~Hee Song, Jiaman Wu, Clayton Washington, Brian~M Sadler, Wei-Lun Chao, and Yu~Su. 2022.
\newblock \href {https://arxiv.org/abs/2212.04088} {Llm-planner: Few-shot grounded planning for embodied agents with large language models}.
\newblock \emph{arXiv preprint arXiv:2212.04088}.

\bibitem[{Su et~al.(2023)Su, Shi, Kasai, Wang, Hu, Ostendorf, Yih, Smith, Zettlemoyer, and Yu}]{sy-etal-2023-oneemb}
Hongjin Su, Weijia Shi, Jungo Kasai, Yizhong Wang, Yushi Hu, Mari Ostendorf, Wen-tau Yih, Noah~A. Smith, Luke Zettlemoyer, and Tao Yu. 2023.
\newblock \href {https://arxiv.org/abs/2212.09741} {One embedder, any task: Instruction-finetuned text embeddings}.
\newblock In \emph{Proc. of the ACL}.

\bibitem[{Su(2023)}]{kexuefm-9529}
Jianlin Su. 2023.
\newblock \href {https://kexue.fm/archives/9529} {Why are all llms now decoder-only architectures?}

\bibitem[{Svajlenko et~al.(2014)Svajlenko, Islam, Keivanloo, Roy, and Mia}]{svajlenko-2014-bcb}
Jeffrey Svajlenko, Judith~F. Islam, Iman Keivanloo, Chanchal~K. Roy, and Mohammad~Mamun Mia. 2014.
\newblock \href {https://doi.org/10.1109/ICSME.2014.77} {Towards a big data curated benchmark of inter-project code clones}.
\newblock In \emph{2014 IEEE International Conference on Software Maintenance and Evolution}, pages 476--480.

\bibitem[{Wang et~al.(2023)Wang, Xie, Jiang, Mandlekar, Xiao, Zhu, Fan, and Anandkumar}]{wang2023voyager}
Guanzhi Wang, Yuqi Xie, Yunfan Jiang, Ajay Mandlekar, Chaowei Xiao, Yuke Zhu, Linxi Fan, and Anima Anandkumar. 2023.
\newblock \href {https://arxiv.org/abs/2305.16291} {Voyager: An open-ended embodied agent with large language models}.
\newblock \emph{arXiv preprint arXiv:2305.16291}.

\bibitem[{Wang et~al.(2022{\natexlab{a}})Wang, Yang, Huang, Jiao, Yang, Jiang, Majumder, and Wei}]{wang2022text}
Liang Wang, Nan Yang, Xiaolong Huang, Binxing Jiao, Linjun Yang, Daxin Jiang, Rangan Majumder, and Furu Wei. 2022{\natexlab{a}}.
\newblock \href {https://arxiv.org/abs/2212.03533} {Text embeddings by weakly-supervised contrastive pre-training}.
\newblock \emph{arXiv preprint arXiv:2212.03533}.

\bibitem[{Wang et~al.(2024)Wang, Yang, Huang, Yang, Majumder, and Wei}]{wang2024improving}
Liang Wang, Nan Yang, Xiaolong Huang, Linjun Yang, Rangan Majumder, and Furu Wei. 2024.
\newblock Improving text embeddings with large language models.
\newblock \emph{arXiv preprint arXiv:2401.00368}.

\bibitem[{Wang et~al.(2022{\natexlab{b}})Wang, Wu, and Neubig}]{wang-etal-2022-english}
Yaushian Wang, Ashley Wu, and Graham Neubig. 2022{\natexlab{b}}.
\newblock \href {https://aclanthology.org/2022.emnlp-main.621} {{E}nglish contrastive learning can learn universal cross-lingual sentence embeddings}.
\newblock In \emph{Proc. of the EMNLP}, pages 9122--9133, Abu Dhabi, United Arab Emirates.

\bibitem[{Wieting et~al.(2020)Wieting, Neubig, and Berg-Kirkpatrick}]{wieting-etal-2020-bilingual}
John Wieting, Graham Neubig, and Taylor Berg-Kirkpatrick. 2020.
\newblock \href {https://doi.org/10.18653/v1/2020.emnlp-main.122} {A bilingual generative transformer for semantic sentence embedding}.
\newblock In \emph{Proc. of the EMNLP}, pages 1581--1594, Online.

\bibitem[{Williams et~al.(2018)Williams, Nangia, and Bowman}]{williams-etal-2018-broad}
Adina Williams, Nikita Nangia, and Samuel Bowman. 2018.
\newblock \href {https://doi.org/10.18653/v1/N18-1101} {A broad-coverage challenge corpus for sentence understanding through inference}.
\newblock In \emph{Proc. of the NAACL-HLT}, pages 1112--1122, New Orleans, Louisiana.

\bibitem[{Xiao et~al.(2023)Xiao, Liu, Zhang, and Muennighof}]{xiao2023c}
Shitao Xiao, Zheng Liu, Peitian Zhang, and Niklas Muennighof. 2023.
\newblock C-pack: Packaged resources to advance general chinese embedding.
\newblock \emph{arXiv preprint arXiv:2309.07597}.

\bibitem[{Yan et~al.(2021)Yan, Li, Wang, Zhang, Wu, and Xu}]{yan-etal-2021-consert}
Yuanmeng Yan, Rumei Li, Sirui Wang, Fuzheng Zhang, Wei Wu, and Weiran Xu. 2021.
\newblock \href {https://doi.org/10.18653/v1/2021.acl-long.393} {{C}on{SERT}: A contrastive framework for self-supervised sentence representation transfer}.
\newblock In \emph{Proc. of the ACL-IJCNLP}, pages 5065--5075, Online.

\bibitem[{Yang et~al.(2020)Yang, Cer, Ahmad, Guo, Law, Constant, Hernandez~Abrego, Yuan, Tar, Sung et~al.}]{yang-etal-2020-multilingual}
Yinfei Yang, Daniel Cer, Amin Ahmad, Mandy Guo, Jax Law, Noah Constant, Gustavo Hernandez~Abrego, Steve Yuan, Chris Tar, Yun-hsuan Sung, and 1 others. 2020.
\newblock \href {https://doi.org/10.18653/v1/2020.acl-demos.12} {Multilingual universal sentence encoder for semantic retrieval}.
\newblock In \emph{Proc. of the ACL: System Demonstrations}, pages 87--94, Online.

\bibitem[{Yang et~al.(2021)Yang, Yang, Cer, Law, and Darve}]{yang-etal-2021-universal}
Ziyi Yang, Yinfei Yang, Daniel Cer, Jax Law, and Eric Darve. 2021.
\newblock \href {https://doi.org/10.18653/v1/2021.emnlp-main.502} {Universal sentence representation learning with conditional masked language model}.
\newblock In \emph{Proc. of the EMNLP}, pages 6216--6228, Online and Punta Cana, Dominican Republic.

\bibitem[{Zhang et~al.(2022)Zhang, Thakur, Ogundepo, Kamalloo, Alfonso-Hermelo, Li, Liu, Rezagholizadeh, and Lin}]{zhang2022making}
Xinyu Zhang, Nandan Thakur, Odunayo Ogundepo, Ehsan Kamalloo, David Alfonso-Hermelo, Xiaoguang Li, Qun Liu, Mehdi Rezagholizadeh, and Jimmy Lin. 2022.
\newblock \href {https://arxiv.org/abs/2210.09984} {Making a miracl: Multilingual information retrieval across a continuum of languages}.
\newblock \emph{arXiv preprint arXiv:2210.09984}.

\bibitem[{Zhang et~al.(2021)Zhang, He, Liu, Bing, and Li}]{zhang-etal-2021-bootstrapped}
Yan Zhang, Ruidan He, Zuozhu Liu, Lidong Bing, and Haizhou Li. 2021.
\newblock \href {https://doi.org/10.18653/v1/2021.acl-long.402} {Bootstrapped unsupervised sentence representation learning}.
\newblock In \emph{Proc. of the ACL-IJCNLP}, pages 5168--5180, Online.

\bibitem[{Zhang et~al.(2020)Zhang, He, Liu, Lim, and Bing}]{zhang-etal-2020-unsupervised}
Yan Zhang, Ruidan He, Zuozhu Liu, Kwan~Hui Lim, and Lidong Bing. 2020.
\newblock \href {https://doi.org/10.18653/v1/2020.emnlp-main.124} {An unsupervised sentence embedding method by mutual information maximization}.
\newblock In \emph{Proc. of the EMNLP}, pages 1601--1610, Online.

\bibitem[{Zhao and Huang(2018)}]{zhao2018deepsim}
Gang Zhao and Jeff Huang. 2018.
\newblock \href {https://doi.org/10.1145/3236024.3236068} {Deepsim: deep learning code functional similarity}.
\newblock In \emph{Proc. of the ESEC/FSE}, page 141–151, New York, NY, USA.

\bibitem[{Zweigenbaum et~al.(2016)Zweigenbaum, Sharoff, and Rapp}]{zweigenbaum2016towards}
Pierre Zweigenbaum, Serge Sharoff, and Reinhard Rapp. 2016.
\newblock Towards preparation of the second bucc shared task: Detecting parallel sentences in comparable corpora.
\newblock In \emph{Proceedings of the Ninth Workshop on Building and Using Comparable Corpora}, pages 38--43, Portoroz, Slovenia.

\end{thebibliography}
